\documentclass{article}

\usepackage[preprint]{neurips_2023}

\usepackage[utf8]{inputenc} %
\usepackage[T1]{fontenc}    %
\usepackage{url}            %
\usepackage{booktabs}       %
\usepackage{amsfonts}       %
\usepackage{nicefrac}       %
\usepackage{xcolor}         %

\usepackage{wrapfig}

\usepackage{natbib}
\usepackage{amssymb}
\usepackage{wrapfig}
\usepackage{subfigure}
\usepackage{caption}
\usepackage{graphicx}
\usepackage{enumitem}
\usepackage{amsmath}

\usepackage{tikz-cd}

\usepackage[colorlinks=true,citecolor=brown,linkcolor=brown,urlcolor=brown]{hyperref}

\usepackage{authblk}
\usepackage{colortbl}
\usepackage{algorithm}
\usepackage[noend]{algorithmic}
\usepackage{bbding}
\usepackage{fancyvrb}
\usepackage{listings}
\definecolor{codegreen}{rgb}{0,0.5,0}
\definecolor{codered}{rgb}{0.7,0.1,0.1}
\definecolor{codegray}{rgb}{0.5,0.5,0.5}
\definecolor{codepurple}{rgb}{0.58,0,0.82}
\definecolor{backcolour}{rgb}{0.95,0.95,0.95}
\lstdefinestyle{python}{
    language=Python,
    basicstyle=\ttfamily\scriptsize,
    backgroundcolor=\color{backcolour},   
    commentstyle=\color{codegreen}\ttfamily\slshape,
    keywordstyle=\color{codered}\bfseries,
    numberstyle=\tiny\color{codegray},
    stringstyle=\color{codepurple},    
    xleftmargin=1em,
    framexleftmargin=0.3em,
    breakatwhitespace=false,         
    breaklines=true,                 
    captionpos=b,  %
    keepspaces=true,                 
    numbers=left,                    
    numbersep=4pt,                  
    showspaces=false,                
    showstringspaces=false,
    showtabs=true,                  
    tabsize=1,
    fancyvrb=true
}
 \usepackage[textsize=tiny,disable]{todonotes}

 \newcounter{todocounter}

\newcommand{\todozlf}[2][]{\todo[color=yellow!80!black, #1]{LF: #2}}
\newcommand{\todoilzlf}[2][]{\todo[inline, size=\small, color=yellow!80!black, #1]{LF: #2}}

\usepackage{amsmath,amsfonts,bm}

\def\eqref#1{equation~\ref{#1}}

\def\1{\bm{1}}

\def\va{{\bm{a}}}

\def\vs{{\bm{s}}}

\def\vz{{\bm{z}}}

\DeclareMathAlphabet{\mathsfit}{\encodingdefault}{\sfdefault}{m}{sl}
\SetMathAlphabet{\mathsfit}{bold}{\encodingdefault}{\sfdefault}{bx}{n}

\def\gA{{\mathcal{A}}}
\def\gB{{\mathcal{B}}}

\def\gE{{\mathcal{E}}}

\def\gS{{\mathcal{S}}}
\def\gT{{\mathcal{T}}}

\def\gV{{\mathcal{V}}}

\def\gX{{\mathcal{X}}}

\def\gZ{{\mathcal{Z}}}

\def\sN{{\mathbb{N}}}

\def\sR{{\mathbb{R}}}

\def\sZ{{\mathbb{Z}}}

\newcommand{\note}[1]{{\color{orange} [Note: #1]}}
\newcommand{\edit}[1]{{\color{black} #1}}

\newcommand{\editneurips}[1]{{\color{blue} #1}}

\newcommand{\ethree}{{\mathrm{E}(3)}}
\newcommand{\sethree}{{\mathrm{SE}(3)}}
\newcommand{\etwo}{{\mathrm{E}(2)}}
\newcommand{\setwo}{{\mathrm{SE}(2)}}
\newcommand{\sothree}{{\mathrm{SO}(3)}}
\newcommand{\sotwo}{{\mathrm{SO}(2)}}
\newcommand{\otwo}{{\mathrm{O}(2)}}
\newcommand{\ztwo}{{\mathbb{Z}^2}}
\newcommand{\rtwo}{{\mathbb{R}^2}}
\newcommand{\rthree}{{\mathbb{R}^3}}
\DeclareMathOperator{\Hom}{Hom}

\newcommand{\base}{{\gB}}

\newtheorem{definition}{Definition}
\newtheorem{example}{Example}  %
\newtheorem{theorem}{Theorem}

\title{Can Euclidean Symmetry be Leveraged\\ in Reinforcement Learning and Planning?}

\author[1]{Linfeng Zhao\thanks{Correspondence: zhao.linf@northeastern.edu.} }
\author[2]{Owen Howell}
\author[1]{Jung Yeon Park}
\author[1]{Xupeng Zhu} 
\author[1]{\\ Robin Walters}
\author[1]{Lawson L.S. Wong}
\affil[1]{ Khoury College of Computer Sciences, Northeastern University, Boston MA, 02115 }
\affil[2]{ Department of Electrical and Computer Engineering, Northeastern University, Boston MA, 02115 }

\begin{document}

\maketitle

\begin{abstract}

In robotic tasks, changes in reference frames typically do not influence the underlying physical properties of the system, which has been known as invariance of physical laws.
These changes, which preserve distance, encompass isometric transformations such as translations, rotations, and reflections, collectively known as the Euclidean group. 
In this work, we delve into the design of improved learning algorithms for reinforcement learning and planning tasks that possess Euclidean group symmetry.
We put forth a theory on that unify prior work on discrete and continuous symmetry in reinforcement learning, planning, and optimal control.
Algorithm side, we further extend the 2D path planning with value-based planning to continuous MDPs and propose a pipeline for constructing equivariant sampling-based planning algorithms. Our work is substantiated with empirical evidence and illustrated through examples that explain the benefits of equivariance to Euclidean symmetry in tackling natural control problems.
\end{abstract}

\addtocontents{toc}{\protect\setcounter{tocdepth}{-1}}

 \section{Introduction}

Robot decision-making tasks often involve the movement of robots in two or three-dimensional Euclidean space. 
\emph{Different reference frames} can be used to model the robot and environment, while the \emph{underlying physics} of the system must be \emph{independent} of the choice of reference frame \citep{Einstein_1905,Wald_1984}.
In this work, we focus on the set of distance preserving transformations between reference frames in $\mathbb{R}^{d}$ forms the Euclidean group $\mathrm{E}(d)$. 
\todozlf{should introduce these later?}
The use of symmetry in decision-making has been studied in model-free or model-based reinforcement learning (RL), planning, optimal control, and other related fields \citep{ravindran2004algebraic,zinkevich_symmetry_2001,van2020mdp,mondal_group_2020,wang_mathrmso2-equivariant_2021,zhao_integrating_2022}. Despite this, there is no unified theory of how symmetry can be utilized to develop better RL or planning algorithms for robotics applications. 
We specifically design algorithms that are equivariant to the Euclidean symmetry groups (and subgroups of the Euclidean group), such as 3D rotations $\sethree$.
\todo{continue editing}

In robotics, we are generally interested in Markov Decision Process (MDP) that describes a robot moving in two or three-dimensional space. 
Motivated by the study of \textit{geometric graphs} in geometric deep learning \citep{bronstein_geometric_2021}, we define \textit{Geometric MDPs} as the class of MDPs that correspond to the decision process of a robot moving in Euclidean space.
A question that we aim to answer is: \textit{Can Euclidean symmetry guarantee benefits in (model-based) RL algorithms?} 
We focus on the case when the symmetry group is continuous because it provides additional structure, and consider discrete case as restricted to discrete subgroups.

We present a theoretical framework that studies the linearized dynamics of geometric MDPs and show that the matrices that appear in linearized dynamics are $G$-steerable kernels \citep{cohen_steerable_2016}.
\todo{potentially add: Our work is an extension of prior work where the group action can be continuous. By restricting $G$ to subgroups (such as reflections or discrete rotations), our work reproduces previously reported results.}
Using recent results on parameterizations of steerable kernels \citep{lang_wigner-eckart_2020}, we show that the steerable kernel solution significantly reduces the number of parameters needed to specify the linearized dynamics. In the idealized non-discretized case, the parameter reduction is infinite. This result is especially useful for tasks involving non-linear systems with unknown dynamics.

Lastly, we propose a sampling-based model-based RL algorithm for Geometric MDPs due to continuous spaces.
It extends the prior work from (1) planning on 2D grid with value-based planning \citep{zhao_integrating_2022} and (2) model-free equivariant RL \citep{van_der_pol_mdp_2020,wang_mathrmso2-equivariant_2021} to continuous state and action space. We take inspiration from geometric deep learning \citep{bronstein_geometric_2021} and consider the features in neural networks to transform under Euclidean symmetry. Our algorithm is constructed under to be equivariant with respect to changes of the reference frame, which is usually known beforehand.
We theoretically analyze the reduction in learnable parameters for a variety of tasks and evaluate empirical performance on several RL benchmarks with 2D or 3D rotation and reflection symmetry.
The results show the effectiveness of symmetry and validate the improvement from theory.
Our contributions are summarized as follows. 
\begin{itemize}[leftmargin=*]
    \item We study MDPs that correspond to the movement of a physical agent in two or three dimensional space. We define this class of MDPs as Geometric MDPs and study their symmetric structure.
    \todo{how do we phrase - the structure we are studying?}
    \item By analyzing the linearization of Geometric MDPs to LQR, our theory shows a reduction in the number of free parameters in the ground-truth linearized dynamics and optimal control policy.
    \item Motivated by our theory, we propose a sampling-based model-based RL algorithm that leverages Euclidean symmetry for Geometric MDPs.
    \item Our empirical results demonstrate the effectiveness of our method in solving MDPs on control tasks with continuous symmetries.
    \todo{why we want theory of LQR?}
\end{itemize}

\section{Problem Statement: Symmetry and Choice of Reference Frame}
\label{sec:statement-gmdp}

\begin{figure*}[t]
\centering
\subfigure{
\includegraphics[width=.85\linewidth]{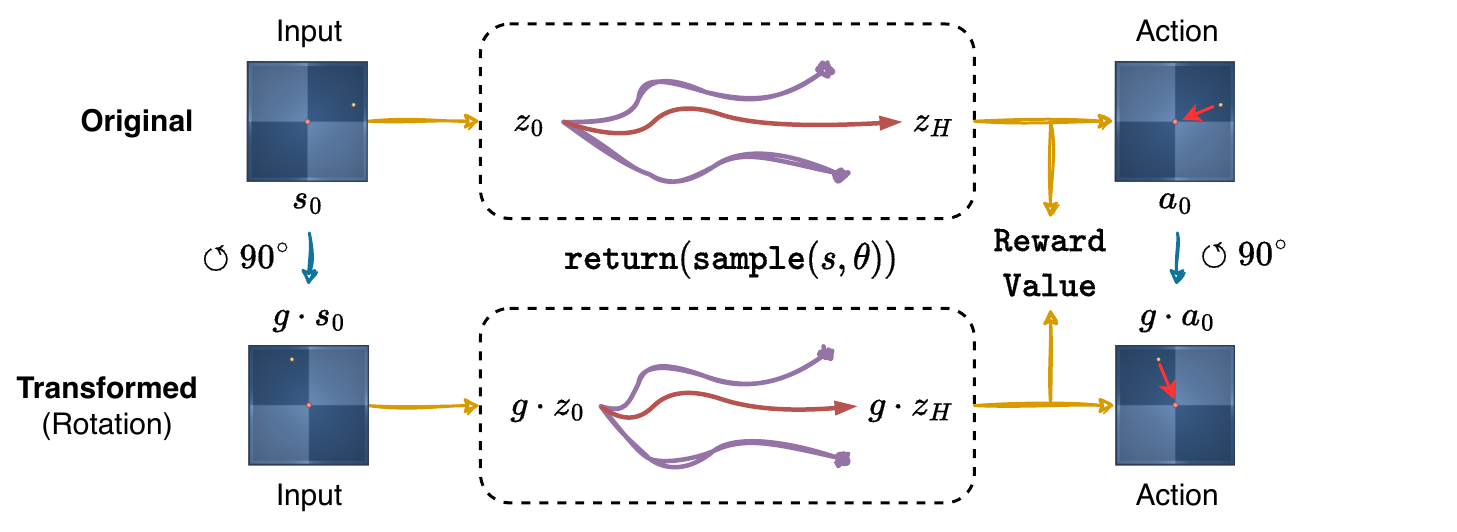}
}
 \vspace{-10pt}
\centering
\caption{
\small
Illustration of equivariance in the proposed sampling-based planning algorithm $a_0 = \texttt{plan} = s_0$. The procedure is equivariant and the learned functions are $G$-equivariant networks.
The sampling procedure produces equivariant trajectories, while the predicted quantities are $G$-invariant, such as values and rewards.
}
\vspace{-14pt}
\label{fig:algo-equiv-sampling}
\end{figure*}

Our focus lies on the potential sharing in geometric transformations of MDPs between reference frames, which are isometric and form the Euclidean symmetry group $\mathrm{E}(d)$ \citep{bronstein_geometric_2021,weiler_general_2021,lang_wigner-eckart_2020}. A key property of this group is that the composition of two changes of reference frames results in another change of reference. These transformations can be expressed in semi-direct product form as $\left(\sR^d,+\right) \rtimes G$, where $G$ is the stabilizer group of origin and the action on a vector $x$ includes a translation part $t$ and rotations/reflections part $g$, i.e., $x \mapsto(t g) \cdot x:=g x+t$ \citep{lang_wigner-eckart_2020}. We concentrate on the compact group $G$, and translations can be implemented by relative position or careful choice of the coordinate system \citep{brandstetter_geometric_2021}.

To transform an MDP (to a different reference frame), we require the MDP to have the group $G$ acting on a set $X$, such as the state or action space \citep{zhao_integrating_2022}. This extends prior work and is analogous to geometric graphs in geometric deep learning for supervised learning \citep{bronstein_geometric_2021}. This definition unifies different types of prior work and allows $X$ to be a homogeneous space, a group, or any other space as long as equipped with a $G$-action \citep{van_der_pol_mdp_2020,wang_mathrmso2-equivariant_2021,zhao_integrating_2022,teng_error-state_2023}. The compact group $G \leq \operatorname{GL}(d)$ can be any group, including the group of proper 3D transformations $\sothree$ or finite subgroups like the icosahedral group or cyclic groups \citep{brandstetter_geometric_2021}.

We define a class of MDPs for which we can study their geometric structure and build the theory section upon \citep{zhao_integrating_2022}. If the group action $\cdot_G : G \times X \to X$ is also continuous, there is an interesting geometric interpretation based on fiber bundle theory \citep{Zee_2016}. This requirement is not mandatory but allows for more rigorous results. Continuous symmetries correspond to conservation laws, while discrete (non-differentiable) symmetries do not have corresponding conservation laws \citep{Zee_2016}. The linearized dynamics of a system are much more constrained when the symmetry of the system is continuous. This is explained in more detail in the theory section.
\todo{Owen: can you check my merged version}
\todo{This looks fine}

\begin{figure*}[t]
\centering
\subfigure{
\includegraphics[width=.85\linewidth]{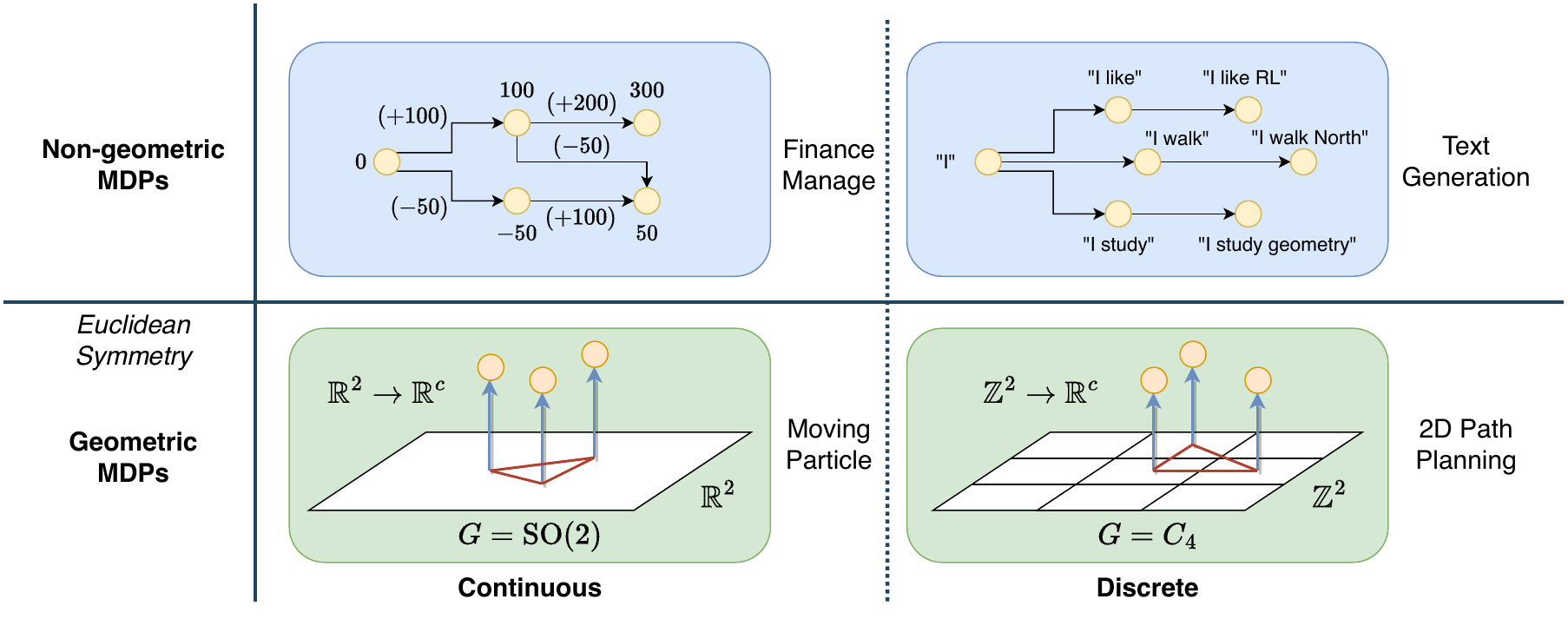}
}
\vspace{-10pt}
\centering
\caption{
\small
Illustration on MDPs with or without underlying geometric structures.
The geometric structures underlying MDPs distinguish the tasks, no matter the underlying space is continuous or discrete.
}
\vspace{-12pt}
\label{fig:geometric-structure-intro}
\end{figure*}

\begin{definition}[Geometric MDP] \label{def:gmdp}
A Geometric MDP (GMDP) $\mathcal{M}$ is an MDP with a (compact) symmetry group $G \leq \operatorname{GL}(d)$ that acts on the state and action space.
It is written as a tuple $\langle \gS, \gA, P, R, \gamma, G, \rho_\mathcal{S}, \rho_\mathcal{A} \rangle$.
The state and action spaces $\gS, \gA$ have (continuous) group actions that transform them, defined by $\rho_\gS$ and $\rho_\gA$.
\end{definition}

For compact Lie groups $G$, if the GMDP has infinitesimal group actions acting on the state and action spaces, there exists a map $p: \gS \times \gA \mapsto \gB$ that projects the state-action space to a lower dimensional base space $\gB$ \citep{cohen_general_2020}. The existence and smoothness of the projection $p$ can be established using principal bundle theory. This is discussed in \ref{sec:theory-appendix}.
We can also \textit{restrict} continuous groups to discrete subgroups for discretized tasks. 
For example, cyclic groups $C_n, n \in \sN$ are subgroups of $\sotwo$.

\subsection{Related Topics}

We discuss geometric graphs and the use of symmetry in reinforcement learning. For further discussion, please see Section~\ref{sec:add_discussion}.

\textbf{Geometric graphs.}
Our definition of GMDP is closely related to the concept of \textit{geometric graphs} \citep{bronstein_geometric_2021,brandstetter_geometric_2021}, which model MDPs as state-action connectivity graphs. Previous works studied algorithmic alignments and dynamic programming on these graphs \citep{xu_what_2019,dudzik_graph_2022}. We propose extending this concept to include additional geometric structures by embedding the MDP into a geometric space such as $\rtwo$ or $\rthree$. In discussion of GMDP, we focus on the most common cases of 2D and 3D Euclidean symmetry \citep{lang_wigner-eckart_2020,brandstetter_geometric_2021,weiler_general_2021}, with the corresponding symmetry groups of $\etwo$ and $\ethree$, respectively. The relation between equivariant message passing and dynamic programming/value iteration on geometric MDPs is discussed in a later section.

\textbf{Symmetry in MDPs and control.}
Symmetry in decision-making has been explored in previous works on MDPs and control, with research on symmetry in MDPs with no function approximation \citep{ravindran2004algebraic,ravindran_symmetries_nodate,zinkevich_symmetry_2001} and symmetry in model-free (deep) RL using equivariant policy networks \citep{van2020mdp,mondal_group_2020,wang_mathrmso2-equivariant_2021}. Besides, symmetry in value-based planning on a 2D grid is analyzed by \citet{zhao_integrating_2022}. We extend this line of work by focusing on MDPs with continuous state and action spaces and sampling-based planning/control algorithms.

The \textit{symmetry properties} in MDPs are specified by equivariance of the transition and reward functions, studied in \citet{zinkevich_symmetry_2001,ravindran2004algebraic,van2020mdp,wang_mathrmso2-equivariant_2021,zhao_integrating_2022}:
\begin{align} \label{eq:symmetry-mdp}
{P}(s' \mid s, a ) & = {P}(g \cdot s' \mid g \cdot s, g \cdot a ) & & \forall g \in G, \forall s, a, s'  \\
	{R}(s, a ) & = {R}(g \cdot s, g \cdot a ) & & \forall g \in G, \forall s, a 
\end{align}

\begin{table}[]
\centering
\caption{Examples of tasks under geometric MDPs. We can quantitatively measure the saving of equivariance.
"Images" refers to panoramic egocentric images $\textbf{$\ztwo\to\sR^{H\times W \times 3}$}$.
$\circ$ denotes group element composition.
We list the quotient space $\gS/G$ to give intuition on saving.
$Gx = \{ g \cdot x \mid g \in G \}$ column shows the $G$-orbit space of $\gS$ (what it is isomorphic to).
}
\small
\begin{tabular}{@{}llllllllll@{}}

\toprule
ID & $G$        & $\gS$                                       & $\gA$      & $\gS / G$                 & $Gx$ & Task                                                   \\ \midrule
1 & $C_4$      & $\ztwo$                                     & $C_4$      & $\ztwo/C_4$          & $C_4$        & 2D Path Planning \citep{tamar_value_2016} \\
2 &$C_4$      & Images & $C_4$      & $\ztwo/C_4$          & $C_4$        & 2D Visual Navigation  \citep{zhao_integrating_2022}                                     \\ \midrule
3 &$\sotwo$   & $\rtwo$                                     & $\rtwo$    & $\sR^+$              & $S^1$    & 2D Continuous Navigation                                           \\
4 &$\sothree$ & $\rthree \times \rthree$                    & $\rthree$  & $\sR^+ \times \sR^3$ & $S^2$    & 3D Free particle (with velocity)                                                 \\
5 &$\sothree$ & $\rthree \rtimes \sothree$                    & $\rthree \times \rthree$  & $\sR^+ \times \sR^3$ & $S^2$    & Moving 3D Rigid Body                                                 \\
6 &$\sotwo$ & $\sotwo$                                  & $\rtwo$ & \textbf{$\{ e \}$}   & $S^1$    & Free Particle on $\sotwo \cong S^1$ manifold                               \\
7 &$\sothree$ & $\sothree$                                  & $\rthree$ & \textbf{$\{ e \}$}   & $S^2$    & Free Particle on $\sothree$ \citep{teng_error-state_2023}                              \\
8 &$\sotwo$   & $\setwo$                                    & $\setwo$   & $\sR^2$              & $S^1$    & Top-down grasping   \citep{zhu_sample_2022}                         \\ 
9 &$\sotwo$   & $(S^1)^2 \times (\rtwo)^2 $                                    & $\sR ^ 2$   &   $S^1 \times (\rtwo)^2$            & $S^1$    & Two-arm manipulation  \citep{tassa_deepmind_2018}    \\ 
\bottomrule

\end{tabular}
\label{tab:examples-geometric-mdps} %
\vspace{-10pt}
\end{table}

\subsection{Illustration and Examples}
\label{subsec:examples-g-mdps}

In Figure~\ref{fig:geometric-structure-intro}, we present visual examples of Geometric MDPs and non-geometric MDPs in both discrete and continuous cases. 
Geometric MDP examples include moving a point robot in a 2D continuous space ($\sR^2$, Example 3 in Table~\ref{tab:examples-geometric-mdps}) or a discrete space ($\sZ^2$, Example 1 \citep{tamar_value_2016}), which is the abstraction of 2D discrete or continuous navigation. 
Table~\ref{tab:examples-geometric-mdps} includes more relevant examples.
We use visual navigation over a 2D grid ($\sZ^2 \rtimes C_4$, Example 2 \citep{lee_gated_2018, zhao_integrating_2022}) as another example of a Geometric MDP.
In this example, each position in $\sZ^2$ and orientation in $C_4$ has an image in $\sR^{H \times W \times 3}$, which is a \textit{feature map} $\sZ^2 \rtimes C_4 \to \sR^{H \times W \times 3}$. %
The agent only navigates on the 2D grid $\sZ^2$ (potentially with an orientation of $C_4$), but not the raw pixel space. %
Example (4) extends to the continuous 3D space and also include linear velocity $\rthree$. 
Alternatively, we can consider (5) moving a rigid body with $\sothree$ rotation.
In (6) and (7), we consider moving free particle positions on $\sotwo,\sothree$, which are examples of optimal control \textit{on manifold} in \citep{lu_-manifold_2023,teng_error-state_2023}.
Here, $G=\sothree$ acts on $\gS=\sothree$ by group composition.
(8) top-down grasping needs to predict $\setwo$ action on grasping an object on plane with $\setwo$ pose. It additionally has translation symmetry, so the state space is technically $\setwo/\setwo=\{ e \}$.
(9) is the \texttt{Reacher} task that we studies later, which controls a two-joint arm.
It is easy to see that because two links are connected, kinematic constraints come in and equivariance does not save so much.
\edit{Additionally, Example (3) is later implemented as \texttt{PointMass}, which is (8) top-down grasping without $\sotwo$ rotation.}

\section{Theory: Why is Symmetry Useful in Geometric MDPs?}

\subsection{Properties of Geometric MDPs}

We start with generic properties of Geometric MDPs, which do not require continuous group actions.
We first consider the equivariance of Bellman operator, which is the foundation of dynamic programming and RL.
This guides us to incorporate symmetry into algorithms on solving Geometric MDPs by enforcing equivariance, according to Theorem 5.1 in \citep{zhao_integrating_2022}.

\begin{theorem}
The Bellman operator of a GMDP is equivariant under Euclidean group $\mathrm{E}(d)$.
\end{theorem}

\begin{theorem}
For a GMDP, value iteration is an $\mathrm{E}(d)$-equivariant geometric message passing.
\end{theorem}

\editneurips{We provide proofs and derivation in Section~\ref{sec:theory-appendix}.}
Due to the relation between GMDP and geometric graphs, we build connection with geometric message passing, which can implement value iteration with $G$-steerable MLP.
This is an extension to \citet{zhao_integrating_2022} Theorem 5.2 on 2D grid, that value iteration is equivariant under discrete subgroups of Euclidean group $\mathrm{E}(2)$: translations, rotations, and reflections. %
We generalize it to groups in the form of $\left(\sR^d,+\right) \rtimes G$, where $G$ is compact.
For translation part, one may use relative/normalized position or induced representations \citep{cohen_general_2020,lang_wigner-eckart_2020}.
For a deeper discussion on non-geometric graphs, we refer the reader to \citep{dudzik_graph_2022}, which shows the equivalence between dynamic programming on a (general non-geometric) MDP and message passing GNN.

\begin{figure*}[t]
\centering
\subfigure{
\includegraphics[width=.8\linewidth]{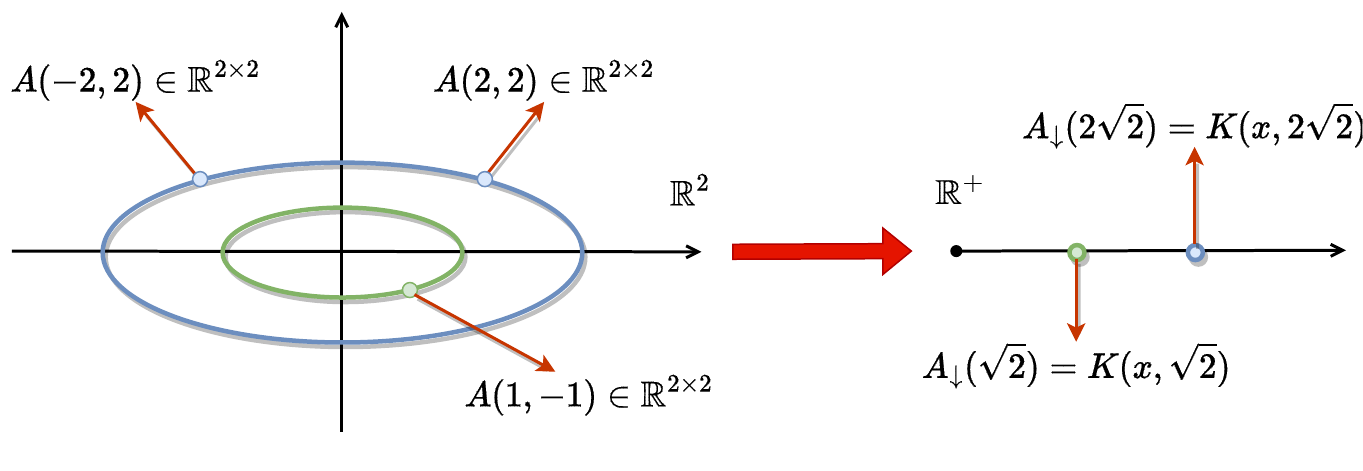}
}
\vspace{-15pt}
\centering
\caption{
\small
This demonstrates how a matrix-valued kernel $A : X \rightarrow \mathbb{R}^{2 \times 2 }$ is \textit{constrained} by the $\sotwo$-steerable kernel constraints on a set of orbits $A( g \cdot p ) = \rho_\text{out}(g) A( p ) \rho_\text{in}(g^{-1})$. 
On the right side, the constraints can be \textit{solved}: the matrices on same orbits (same colors) are related and form a vector space.
Thus, these matrices can be spanned on a basis (\edit{denoted by $K$}) and live in a smaller ``base'' space $\gB = X / G$ with a certain form $A_\downarrow : \gB \rightarrow \mathbb{R}^{2 \times 2}$, potentially with less parameters.
\edit{Further explained in Section \ref{sec:theory-appendix}.}
}
\vspace{-10pt}
\label{fig:demo-kernel}
\end{figure*}

\subsection{Linearizing Geometric MDPs: $G$-steerable Kernel Constraints}

Geometric MDPs can model any discrete-time, stochastic, nonlinear dynamical system.
However, the functions of interest in GMDPs are generally nonlinear, which is challenging to analyze using tools from representation theory.
In this subsection, we derive the iterative linearization of dynamics of GMDPs to get $G$-equivariant linear maps.
There are two reasons: \textit{(1)} if infinitesimal group actions on state-action space exists, the symmetry of the nonlinear GMDP is \textit{equivalent} to $G$-steerable constraints of the linear dynamics, \textit{(2)} the linearized dynamics is connected to LQR and is easier to analyze, such as the dimensions of the (linear) dynamics function, policy function, and more.

\textbf{Iterative Linearization.}
We assume the dynamics is deterministic $f: \gS \times \gA \to \gS$ and \textit{iteratively linearize} $f$ at each step.
This linearized problem is naturally connected to \textit{time-varying} iterative Linear Quadratic Regulator (iLQR), which assumes deterministic \textit{linear dynamics} and \textit{quadratic cost}.
We highlight the linearization procedure of $f(\vs_t , \va_t )$, where matrices $A$ and $B$ depend \textit{arbitrarily} on time step ${\color{red}t}$. Later, we assume that it only depends on state and action $(\vs_t , \va_t )$.
\begin{equation}
\text{Original:}  \quad   \vs_{t+1} = f(\vs_t , \va_t ) \quad \to \quad
\text{Linearized at step ${\color{red}t}$:} \quad  \vs_{t+1} = A_{\color{red}t}  \cdot  \vs_{t} + B_{\color{red}t}  \cdot \va_{t}
\end{equation}

\begin{theorem}
If a Geometric MDP has an infinitesimal $G$-action on the state-action space $\gS \times \gA$, the linearized dynamics is also $G$-equivariant: the matrix-value functions $A : \mathcal{S} \times \mathcal{A} \rightarrow \mathbb{R}^{d_{\mathcal{S} } \times d_{ \mathcal{S} }}$ and $B : \mathcal{S} \times \mathcal{A} \rightarrow \mathbb{R}^{d_{ \mathcal{S} }\times d_{ \mathcal{A} }}$ satisfy $G$-steerable kernel constraints.
\end{theorem}

\edit{We explain what this theorem implies.} 
	 Under infinitesimal symmetry transformation $g \approx 1_{G} \in G$, let us suppose that the state and action space transform as $\vs \mapsto \rho_\gS(g) \cdot \vs , \va \mapsto \rho_\gA(g) \cdot \va$
	where $\rho_\gS$ and $\rho_\gA$ are representations of the group $G$. Under infinitesimal symmetry transformation $g \approx 1_{G} \in G$, the dynamics must satisfy,
	\begin{align}\label{Equation:Symmetry of Non-linear dynamics in LQR}
    \rho_\gS(g) \cdot f( \vs ,  \va ) = f( \rho_\gS(g)   \cdot \vs , \rho_\gA(g)   \cdot \va )
	\end{align}
    Let us consider the linearized problem at point $p = ( \vs_{0} , \va_{0} )$. Assuming that the state and control do not change too drastically over a short period of time {and the time-varying $A$ and $B$ only depend on the linearization point $p$ but not other factors}, we can approximate the true dynamics as
	\begin{align}
		\vs_{t+1} = A {\color{red}(p)} \cdot  \vs_{t} + B {\color{red}(p)} \cdot \va_{t}, \quad A : \mathcal{S} \times \mathcal{A} \rightarrow \mathbb{R}^{d_{\mathcal{S} } \times d_{ \mathcal{S} }}, \quad B : \mathcal{S} \times \mathcal{A} \rightarrow \mathbb{R}^{d_{ \mathcal{S} }\times d_{ \mathcal{A} }}.
	\end{align}
    Now, linearizing $f$ and using the symmetry constraint, the matrix-valued functions $A{\color{red}(p)}$ and $B{\color{red}(p)}$ must satisfy the constraints
	\begin{align}\label{Equation: A and B kernel}
		\forall g\in G, \quad	A {\color{red} (g \cdot p) } = \rho_\gS(g) A {\color{red}(p)} \rho_\gS(g^{-1}) , \quad B {\color{red} (g \cdot p) } = \rho_\gS(g) B {\color{red}(p)} \rho_\gA(g^{-1}) 
	\end{align}
so that $A$ is a $G$-steerable kernel with input and output representation $\rho_\gS$ and $B$ has input type $\rho_\gS$ and output type $\rho_\gA$.
 The kernel constraints relate $A( p )$ and $B( p )$ at different points.

\begin{table}[t]
\caption{Equivariant dimension reduction for linearized dynamics. 
\edit{This table highlights the reduced dimensions of spaces of kernels.}
}
\small
\centering
\begin{tabular}{l|llllllll}
\toprule
Task                       & $\gS$                     & $\gA$                                    & $G$        & $\rho_\gS$   & $\rho_\gA$   & $ \gX( \gS \times \gA ) $  & $ \gX( \base ) $                                           \\ \midrule
Free Particle in 2D & $\rtwo \times \rtwo $     & $ \rtwo $      & $\sotwo$   & $\rho_{std}$ & $\rho_{std}$ & $ \mathbb{R}^{16} $ & $ \mathbb{R}^{+} \times \mathbb{R}^{14}  $                  \\
Free Particle in 3D     & $\rthree \times \rthree $ & $\rthree$ & $\sothree$ & $\rho_{std}$ & $\rho_{std}$ & $\mathbb{R}^{32}$   & $(\mathbb{R}^{+} )^{2}\times \mathbb{R}^{4+d_{S} + d_{A}} $ \\ \bottomrule
\end{tabular}
\label{tab:examples-lqr-main} %
\end{table}

\textbf{Benefits for control.}
Symmetry further enables better control.
We can further show that the policy and value function can be parameterized with less parameters based on the discrete algebraic Riccati equation (DARE) for time-varying LQR problem.

\begin{theorem}
The LQR feedback matrix in $a_t^\star = - K(p) s_t$ and value matrix in $V = s_t^\top P(p) s_t$ are $G$-steerable kernels, or matrix-valued functions: $K : \mathcal{S} \times \mathcal{A} \rightarrow \mathbb{R}^{d_{\mathcal{A} } \times d_{ \mathcal{S} }}$ and $P : \mathcal{S} \times \mathcal{A} \rightarrow \mathbb{R}^{d_{ \mathcal{S} }\times d_{ \mathcal{S} }}$.
\end{theorem}

\paragraph{What tasks are suitable for Euclidean equivariance?}
We find that the ratio of state features that can transform under Euclidean symmetry (change of reference frame) decides the reduction of space $\gS \times \gA \mapsto \gB$, where the dimension of the space of steerable kernels quantitatively shows this.
This means that tasks with more kinematic constraints are harder to use Euclidean equivariance.
In \edit{summary}: 
(1) More equivariant features (associated with non-trivial representations) lower system dimensions.
(2) For kinematic chains, it inherently has local coordinates, which is hard to use Euclidean symmetry w.r.t. the global reference frame.
It needs to explicitly consider constraints, such as the case in position-based dynamics \citep{tsai_position_2017} or particle-based dynamics \citep{han_learning_2022}.
\edit{We further discuss examples and computation of dimensions in Appendix Section \ref{sec:theory-appendix}.}

\subsection{Examples: Linearized Tasks}

In Table~\ref{tab:examples-lqr-main}, we show a few examples on how we linearize a GMDP problem and compute the dimensions of the space of $G$-steerable kernels with $\gX(\gB)$ or without equivariance $\gX(\gS  \times \gA)$, which gives us a \textit{measure} of hardness of Geometric MDPs.
We further explain in \edit{appendix}.
Here, we give a concrete example on solving the kernel constraints of a 2D particle, which is also the first continuous-space example in Table~\ref{tab:examples-geometric-mdps}. We visualize an even simplified $\rtwo$ case on Figure~\ref{fig:demo-kernel}.

\begin{example}
A 2D free particle with position and velocity has state space $\sR^2 \times \sR^2$ and action space $\sR^2$. %
The state space is transformed by $\sotwo$ under two standard representations $\rho_\gS = \rho_1 \oplus \rho_1 $, and the action transformed under $\rho_\gA = \rho_1$.
\end{example}

The base space is given by $\base = \mathbb{R}^{+} \times \mathbb{R}^{4}$. Using Proposition E.6. in \cite{Lang_2020}, a basis for the steerable kernels of input type $(\rho_{1} , V_{1} )$ and output type $(\rho_{1} , V_{1} )$ is given by
\begin{align*}
K_{11}(x) = c_{1}\begin{bmatrix}
    1 & 0 \\
    0 & 1
\end{bmatrix} + c_{2}\begin{bmatrix}
    0 & -1 \\
    1 & 0
\end{bmatrix} + c_{3} \begin{bmatrix}
    \cos(2x) & \sin(2x) \\
    \sin(2x) & -\cos(2x)
\end{bmatrix} + c_{4} \begin{bmatrix}
    -\sin(2x) & \cos(2x) \\
    \cos(2x) & -\sin(2x)
\end{bmatrix} 
\end{align*}
where each $c_{j} \in \mathbb{R}$. This result was first derived in \cite{Weiler_2018}. 
Then, using the form of linear dynamics, we get:
\begin{align*}
    A( p_{\downarrow} , x ) = \begin{bmatrix}
        K_{A}^{(1,1)}( p_{\downarrow},x) , & K_{A}^{(1,2)}( p_{\downarrow},x) \\
        K_{A}^{(2,1)}( p_{\downarrow},x) , & K_{A}^{(2,2)}( p_{\downarrow},x)
    \end{bmatrix}, \quad    B( p_{\downarrow} , x ) = \begin{bmatrix}
   K_{B}^{(1,1)}( p_{\downarrow},x) \\
   K_{B}^{(2,1)}( p_{\downarrow},x)
    \end{bmatrix}
\end{align*}
where each $K_{A}^{(i,j)}$ and $K_{B}^{(k,l)}$ take the above form.
They have coefficients $c_{kA}^{(i,j)} : \mathbb{R}^{+} \times \mathbb{R}^{4} \rightarrow \mathbb{R} $ and $c_{kB}^{(i,j)} : \mathbb{R}^{+} \times \mathbb{R}^{4} \rightarrow \mathbb{R} $ are \textit{not constrained} by symmetry and can be parameterized by a neural network. If we amalgamate each of the coefficients $c_{kA}^{(i,j)}$ and $c_{kB}^{(i,j)}$ into a single vector output $C$, the system dynamics can be learned by specifying $C : \mathbb{R}^{+} \times \mathbb{R}^{4} \rightarrow \mathbb{R}^{16 + 8}$.
This should be contrasted with the non-equivariant case, where one needs to learn 
$A : \mathbb{R}^{6} \rightarrow \mathbb{R}^{16} \text{ and } B: \mathbb{R}^{6} \rightarrow \mathbb{R}^{8}$.
\edit{The table also shows 3D free particle example, which has more savings. We discuss in detail in Appendix Section \ref{sec:theory-appendix} and also demonstrate numerically.}

\section{Symmetry in Sampling-based Planning Algorithms}

In this section, we aim to exploit the symmetry in Geometric MDPs $G \leq \operatorname{GL}(d)$, such as rotations and reflections, for sampling-based planning.
We extend prior work \citep{zhao_integrating_2022} that uses value-based planning on a discrete state space $\ztwo$ and discrete group $D_4$ to continuous case, necessitating sampling-based planning.
The idea is to ensure that the algorithm $a_t = \texttt{plan}(s_t)$ produces same actions up to transformations, i.e., it is $G$-equivariant: $g \cdot a_t \equiv g \cdot \texttt{plan}(s_t) = \texttt{plan}(g\cdot s_t)$, shown in Figure~\ref{fig:algo-equiv-sampling}.
The principle is potentially applicable for MDPs with other groups.

\textbf{Components.}
We use TD-MPC \citep{hansen_temporal_2022} as the backbone of our implementation and introduce their procedure.
The principle of designing an equivariant sampling-based planning algorithm does not limit to a particular algorithm.

\begin{itemize}[leftmargin=*]
\item \textit{Planning with learned models.} We use MPPI (Model Predictive Path Integral) control method \citep{williams_model_2015,williams_aggressive_2016,williams_model_2017,williams_information_2017}, as also done in \citep{hansen_temporal_2022}.
We sample $N$ trajectories with horizon $H$ using the learned dynamics model with actions from a learned policy, and estimate the expectation of total return.
\item \textit{Training models.} The learnable components in equivariant TD-MPC include: an \texttt{encoder} that processes input observation, \texttt{dynamics} and \texttt{reward} network that simulate the MDP, and \texttt{value} and \texttt{policy} network that guide the planning process.
\item \textit{Loss.} The only requirement is that loss is $G$-invariant. The loss terms in TD-MPC include value prediction MSE loss and dynamics/reward consistency MSE loss, which all satisfy invariance.
\end{itemize}

\textbf{Integrating symmetry.}
\citep{zhao_integrating_2022} consider how the Bellman-Operator transforms under symmetry transformation. For sampling based methods, one needs to consider how the sampling procedure changes under symmetry transformation. Specifically, under a symmetry transformation, different sampled trajectories must transform equivariently. This is show in \ref{fig:algo-equiv-sampling} The equivariance of the transition model in sampling based approches to machine learning has also been studied in \citep{park_learning_2022}.
There are several components that need $G$-equivariance, and we discuss them step-by-step and visualize in Figure~\ref{fig:algo-equiv-sampling}.

\begin{enumerate}[leftmargin=*]
\item \textbf{\texttt{dynamics} and \texttt{reward} model.} In the definition of symmetry in Geometric MDPs (and symmetric MDPs \citep{ravindran2004algebraic,van_der_pol_mdp_2020,zhao_integrating_2022}) in Equation~\ref{eq:symmetry-mdp}, the transition and reward function are $G$-invariant. Therefore, in implementation, the transition network is deterministic and uses a $G$-equivariant MLP, and the reward network is constrained to be $G$-invariant.
\item \textbf{\texttt{value} and \texttt{policy} model.} The optimal value function produces a scalar for each state and is $G$-invariant, and the optimal policy function is $G$-equivariant \citep{ravindran2004algebraic}. Assuming if we use equivariant/invariant transition and reward networks in updating our value function $\mathcal{T}[V_\theta] = \sum_\va R_\theta(\vs,\va) + \gamma \sum_{\vs'} P_\theta(\vs'|\vs, \va) V_\theta(\vs')$, the learned value network $V_\theta$ will also satisfy the symmetry constraint. Similarly, we can extract policy from the value network, which is also equivariant \citep{van_der_pol_mdp_2020,wang_mathrmso2-equivariant_2021,zhao_integrating_2022}.
\item \textbf{\texttt{MPC} procedure.} We consider the equivariance in MPC procedure in three parts: \texttt{sample} trajectories from the MDP, compute \texttt{return} of them, and use gradients of loss to \texttt{update}: $\theta' = \texttt{update}(\texttt{return}(\texttt{sample}(s, \theta)))$. We discuss the equivariance in the next subsection. 
\end{enumerate}

Note that TD-MPC and similar model-based RL algorithms typically plan in a latent space $h: \gS \to \gZ$ by $\bar{f}(\vz, \va) = \vz'$, while we still use $\vs$ for simplicity in notation. In implementation, the difference is merely to also enforce $G$-equivariance in $h: \rho_\gZ(g) \cdot h(\vs) = h(\rho_\gS(g) \cdot \vs)$.

We list the equivariance or invariance conditions that each network needs to satisfy.
Alternatively, for scalar functions, we can also say they transform under \textit{trivial} representation $\rho_0$ and is thus invariant.
\edit{
All modules are implemented via $G$-steerable equivariant MLPs, which means that all variables in the implementation are transformable by $G$.
\todozlf{edit this: how to implement using equivariant MLPs}
}
\begin{align}
f_\theta &: \gS \times \gA \to \gS :& \rho_\gS(g) \cdot f_\theta(\vs_{t}, \va_{t}) &=  f_\theta(\rho_\gS(g) \cdot \vs_{t}, \rho_\gA(g) \cdot \va_{t}) \\
R_\theta &: \gS \times \gA \to \sR :& {R}_\theta(\vs_{t}, \va_{t} ) &=  {R}_\theta(\rho_\gS(g) \cdot \vs_{t}, \rho_\gA(g) \cdot \va_{t}) \\
Q_\theta &: \gS \times \gA \to \sR :& {Q}_\theta(\vs_{t}, \va_{t} ) &=  {Q}_\theta(\rho_\gS(g) \cdot \vs_{t}, \rho_\gA(g) \cdot \va_{t}) \\
\pi_\theta &: \gS \to \gA  :& \rho_\gA(g) \cdot {\pi}_\theta(\cdot \mid \vs_{t}) &=  {\pi}_\theta(\cdot \mid \rho_\gS(g) \cdot \vs_{t})
\end{align}

\textbf{Equivariance in \texttt{MPC}.}
We analyze how to constraint the underlying MPC planner to be equivariant.
We use MPPI (Model Predictive Path Integral) \citep{williams_model_2015,williams_model_2017}, which has been used in TD-MPC for action selection.
An MPPI procedure samples multiple trajectories $\{ \tau_i \}$ and compute returns to estimate a good action.
We use $\texttt{sample}$ to refer to the procedure that samples a trajectory $\tau_i$ using the learned models, which is given by: $\tau_i \equiv \texttt{sample}(g \cdot \vs_0 ; f_\theta, R_\theta, Q_\theta, \pi_\theta) = (\vs_1, \va_1, \vs_2, \va_2, \ldots, \vs_H)$.
A trajectory is transformed element-wise by a transformation $g$: $g\cdot \tau_i = (g \cdot \vs_1, g \cdot \va_1, g \cdot \vs_2, g \cdot \va_2, \ldots, g \cdot \vs_H)$.

The next step is to compute the return of sampled trajectories. 
Under expectation, the return is computed by the next equation.
We can study how the return is transformed.
\begin{align}
	\texttt{return}({\tau}) & = \mathbb{E}_{\tau}\left[\gamma^H Q_\theta\left(\vs_H, \va_H\right)+\sum_{t=0}^{H-1} \gamma^t R_\theta\left(\vs_t, \va_t\right)\right] = \mathbb{E}_{\tau}\left[ U(\vs_{1:H}, \va_{1:H-1}) \right] \\
	\texttt{return}(g \cdot {\tau}) & = \mathbb{E}_{g \cdot \tau}\left[\gamma^H \rho_0(g) \cdot  Q_\theta\left(g \cdot \vs_H, g \cdot \va_H\right)+\sum_{t=0}^{H-1} \gamma^t \rho_0(g)\cdot  R_\theta\left(g \cdot \vs_t, g \cdot \va_t\right)\right] \label{eq:tranformed_traj} \\
	& = \int_{g\in G}\rho_0(g) dg \cdot  \mathbb{E}_{\tau}\left[ U(g \cdot \vs_{1:H}, g \cdot \va_{1:H-1})  \right] \label{eq:extracted_g} \\
		& = \mathbf{1} \cdot  \mathbb{E}_{\tau}\left[ U(\vs_{1:H}, \va_{1:H-1})  \right] = \texttt{return}({\tau}) \label{eq:g_simplify}
\end{align}

In Eq.~\ref{eq:tranformed_traj}, we use $\rho_0(g) = \mathbf  1$ to denote that the output is not affected, so we may extract the term out.
In Eq.~\ref{eq:extracted_g}, $dg$ is a Haar measure that absorbs normalization factor, and we can extract the term from expectation.
Eq.~\ref{eq:g_simplify} uses the invariance of $Q_\theta$ and $R_\theta$.
In other words, the return under the $G$-orbit of trajectories, $[\tau_i]_G = \{ g\cdot \tau_i \mid g \in G\}$, is the same, thus \texttt{return} is $G$-invariant.

In summary, for sampling and computing return, they satisfy the following conditions, indicating that the procedure $\texttt{return}(\texttt{sample}(s, \theta))$ is invariant, i.e. not changed under group transformation for any $g$.
Thus, we do \textbf{not} need to additionally enforce anything for $\texttt{MPC}$ procedure.
We use $\texttt{return}(\tau_i)$ to indicate the return of a specific trajectory $\tau_i$ and $g\cdot \tau_i$ to denote group action on it.
\begin{align}
\texttt{sample} &: \vs_0, \theta \mapsto \tau_i  :& g \cdot \tau_i &\sim \texttt{sample}(g \cdot s_0 ; f_\theta, R_\theta, Q_\theta, \pi_\theta) \\
\texttt{return} &: \tau_i \mapsto \sR  :& \texttt{return}(\tau_i) &= \texttt{return}(g \cdot \tau_i)
\end{align}

\textbf{Remark.}
There are some benefits of explicitly considering symmetry in continuous control.
The possibility of hitting orbits is negligible, so there is no need for orbit-search on symmetric states in forward search in continuous control.
Additionally, the planning algorithm implicitly plans in a smaller continuous MDP $\mathcal{M} / G$ \citep{ravindran2004algebraic}.
Furthermore, from equivariant network literature \citep{elesedy_provably_2021}, the generalization gap for learned equivariant policy and value networks are smaller, which allows them to generalize better.

\section{Evaluation: Sampling-based Planning}

In this section, we present the setup and results for our proposed sampling-based planning algorithm: equivariant version of TD-MPC.
The additional details and results are available in \edit{Section \ref{sec:add_results}}.

\begin{figure*}[t]
\centering
\subfigure{
\includegraphics[width=.9\linewidth]{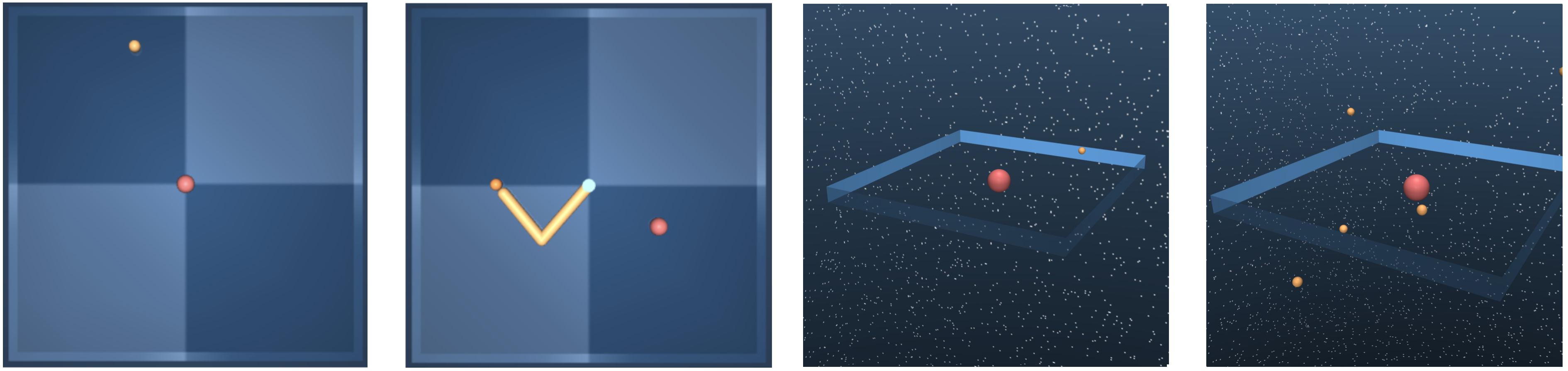}
}
\centering
\caption{
\small
Sampled tasks that we use in experiments:
(1) \texttt{PointMass} in 2D, (2) \texttt{Reacher}, (3) \texttt{PointMass} customized 3D version, and (4) \texttt{PointMass} customized 3D version with multiple particles to control.
}
 \vspace{-10pt}
\label{fig:env-collection}
\end{figure*}

\textbf{Tasks.}
We verify the algorithm on a few selected tasks from DeepMind Control suite (DMC) and several customized ones, visualized in Figure~\ref{fig:env-collection}.
One task is 2D particle moving in $\sR^2$, named \texttt{PointMass}. We customize tasks based on it: (1) 3D particle moving in $\sR^3$ (disabled gravity), and (2) 3D $N$-point moving that has several particles to control simutaneously. The goal is still to move particle(s) to a target position (the origin).
We also experiment two-arm manipulation tasks, \texttt{Reacher} (easy and hard), where the goal is to move the end-effector to a random position in a plane.
\edit{In Section \ref{sec:add_results}, we additionally discuss tasks that Euclidean symmetry do not practically work better, which is related to the ratio of equivariant features discussed in theory.}

\textbf{Experimental setup.}
We compare against the non-equivariant version of TD-MPC \citep{hansen_temporal_2022}.
Here, we by default make all components equivariant as described in the algorithm section.
In \edit{appendix}, we include ablation studies for disabling or enabling each equivariant component.
The training procedure follows TD-MPC \citep{hansen_temporal_2022}. We use the state as input and for equivariant TD-MPC, we divide the orignal hidden dimension by $\sqrt{N}$ where $N$ is the group order to keep the number of parameters roughly equal to the equivariant and non-equivariant version. We mostly follow the original hyperparameters except for \texttt{seed\_steps}. We use $5$ random seeds for each method.

\textbf{Algorithm setup: equivariance.}
We use discretized subgroups in implementing $G$-equivariant MLPs with \texttt{escnn} package \citep{weiler_general_2021}, as they perform more stably and 
For 2D case, we use $\otwo$ subgroups: dihedral groups $D_4$ and $D_8$ ($4$ or $8$ rotation components).
For 3D case, we use Icosahedral group and Octahedral group, which are finite subgroups of $\sothree$ with order $60$ and $24$, respectively.

\begin{figure*}[t]
\centering
\subfigure{
\includegraphics[width=1.\linewidth]{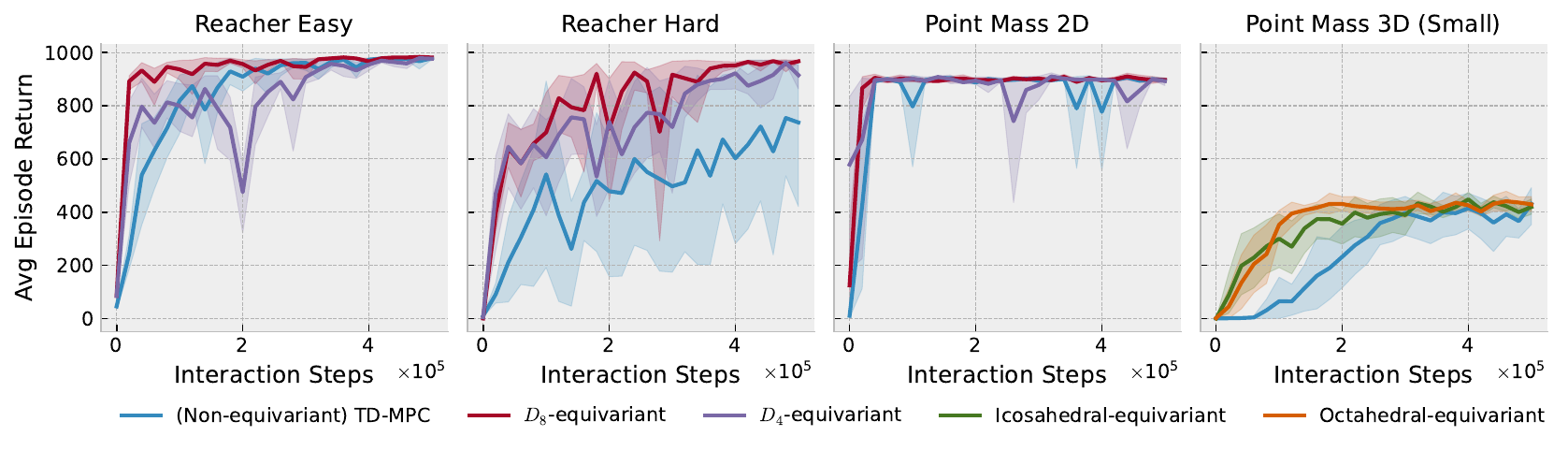}
}
 \vspace{-18pt}
\centering
\caption{
\small
Results on \texttt{Reacher}, default \texttt{PointMass} 2D, and customized 3D \texttt{PointMass} with smaller target.  %
}
\vspace{-10pt}
\label{fig:env-reward-curves-combined}
\end{figure*}

\begin{figure*}[t]
\centering
\subfigure{
\includegraphics[width=1.\linewidth]{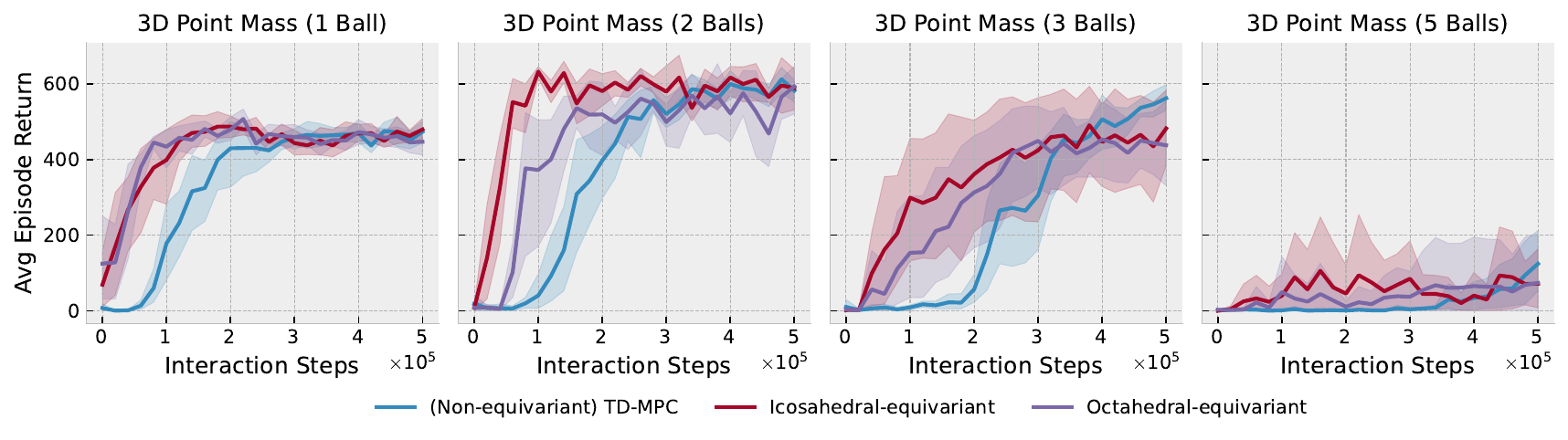}
}
 \vspace{-18pt}
\centering
\caption{
\small
Results on a set of customized 3D $N$-ball \texttt{PointMass} tasks, with $N=1,2,3,5$.
}
\vspace{-14pt}
\label{fig:env-reward-curves-3d-point}
\end{figure*}

\textbf{Results.}
In Figure~\ref{fig:env-reward-curves-combined} and \ref{fig:env-reward-curves-3d-point}, we show the reward curves in evaluation.
\texttt{Reacher} \texttt{easy} and \texttt{hard} are top-down where the goal is to reach a random 2D position.
If we rotate the MDP, the angle between the first and second links is not affect, i.e. $G$-invariant.
The first joint and the target position are transformed under rotation, so we set to $\rho_1$ standard representation (2D rotation matrices).
The complete state and action representations are given in Table~\ref{tab:examples-geometric-mdps}.
The system has $\otwo$ rotation and also reflection symmetry, and we use $D_8$ and $D_4$ groups.
Shown in Figure~\ref{fig:env-reward-curves-combined}, $D_8$ outperforms the non-equivariant TD-MPC by noticable margins, especially on \texttt{hard} one. $D_4$ is slightly worse than $D_8$ but still better than the baseline.
With higher order discrete subgroups, the performance plateaus and does not worth the additional compute.

The default \texttt{PointMass} 2D version seems easy to solve, while $D_8$-equivariant version still learns  faster.
Thus, we design 3D version of \texttt{PointMass} and use $\sothree$ subgroups to implement 3D equivariant version of TD-MPC, because the implementation is significantly easier and the computational cost is lower compared to continuous version, which needs to convert between frequency domain and spatial domain.
Figure~\ref{fig:env-reward-curves-3d-point} shows $N=1,2,3,5$ balls in 3D \texttt{PointMass}, and the rightmost figure in Figure~\ref{fig:env-reward-curves-combined} shows $1$-ball 3D version with smaller target ($0.02$ compared to $0.03$ in $N$-ball version).
We find the Icosahedral (order $60$) equivariant TD-MPC always learns faster and uses less samples to achieve best rewards. Octahedral (order $24$) equivariant version is pretty close and is also mostly better.
The best absolute rewards in 1-ball case is interestingly lower than $2$- and $3$-ball, which may be caused by higher possible reward due to 2 or 3 balls that can reach the goal.

We find TD-MPC is especially sensitive to a hyperparameter \texttt{seed\_steps} that controls the number of warmup trajectories.
In contrast, our equivariant version is robust to it and sometimes learn better with less warmup.
We conjecture that this is related to the end-to-end learning of all components in the model-based RL algorithm (transition, reward, policy, value) with task-specific loss (purely reward-driven). Thus, the efficiency of training all components together matters, especially for sparse-reward goal-reaching tasks, where the equivariant network components start to shine.
In the shown curves, we do not use warmup across non-equivariant and equivariant ones and present additional results in \edit{Section \ref{sec:add_results}}.

\section{Conclusion and Discussion}

In conclusion, our work highlights the usefulness of Euclidean symmetry in (model-based) RL algorithms. We argue that MDPs that appear in robotics problems have additional structure and define this special subclass of MDPs as Geometric MDPs.
We show that the linearized approximation of any Geometric MDPs (LQR) satisfies steerable kernel constraints, which significantly reduces the size of the parameter space and is related to the theory of principal bundles.
We furthermore proposed a sampling-based model-based RL algorithm that leverages Euclidean symmetry for Geometric MDPs. Our algorithm outperforms standard reinforcement learning techniques on a variety of common RL benchmarks.
Our theory and proposed algorithm unveil scenarios where Euclidean symmetry can play an important role in achieving significant savings in the number of parameters, but we also find situations where Euclidean symmetry may not bring practical benefits.
\todo{mention in some case it doesn't really save, e.g. locomotion}
Overall, our findings contribute to a deeper understanding of the use of symmetry in (model-based) RL algorithms and provide insights for future research in this area. Specifically, it may be possible to extend our group to more exotic symmetry groups. 

Our work exists a few limitations. It requires known symmetry group, however, the symmetry can be usually determined by the dimension of the robot space, which is normally 3D.
Additionally, we empirically find locomotion tasks do not exhibit great benefits from Euclidean symmetry. As explained by dimension reduction in our theory, changes of reference frame is global to the system. Thus, for robots that have many connected links, all joints are unchanged under

\newpage

\bibliography{zotero-linfeng, main, refs}

\newpage

\appendix

\tableofcontents

\addtocontents{toc}{\protect\setcounter{tocdepth}{2}}

\listoftodos

\section{Outline}

The appendix is organized as follows: (1) additional discussion, including related work and theoretical background, (2) theory, derivation, and proofs, (3) implementation details and further empirical results, and (4) additional mathematical background.

\section{Additional Discussion}
\label{sec:add_discussion}

\subsection{Discussion: Symmetry in Decision-making}

In this work, we study the Euclidean symmetry $\mathrm{E}(d)$ from geometric transformations between \textit{reference frames}.
This is a specific set of symmetries that an MDP can have -- isometric transformations of Euclidean space $\sR^d$, such as the distance is preserved.
This can be viewed as a special case under the framework of MDP homomorphism, where symmetries relate two different MDPs via MDP \textit{homomorphism} (or more strictly, \textit{isomorphism}).
We refer the readers to \citep{ravindran2004algebraic} for more details.
We also discuss symmetry in other related fields.

Classic planning algorithms and model checking have leveraged the use of symmetry properties, \citep{fox_detection_1999, fox_extending_2002, pochter_exploiting_2011, domshlak_enhanced_nodate, shleyfman_heuristics_2015, holldobler_empirical_2015, sievers_structural_nodate, sievers_theoretical_2019, fiser_operator_2019} as evident from previous research. In particular, \citet{zinkevich_symmetry_2001} demonstrate that the value function of an MDP is invariant when symmetry is present. However, the utilization of symmetries in these algorithms presents a fundamental problem since they involve constructing equivalence classes for symmetric states, which is difficult to maintain and incompatible with differentiable pipelines for representation learning. \citet{narayanamurthy_hardness_2008} prove that maintaining symmetries in trajectory rollout and forward search is intractable (NP-hard). To address the issue, recent research has focused on state abstraction methods such as the coarsest state abstraction that aggregates symmetric states into equivalence classes studied in MDP homomorphisms and bisimulation \citep{ravindran2004algebraic, ferns_metrics_2004, li_towards_2006}. However, the challenge lies in that these methods typically require perfect MDP knowledge and do not scale well due to the complexity of constructing and maintaining abstraction mappings \citep{van2020mdp}. To deal with the difficulties of symmetry in forward search, recent studies have integrated symmetry into reinforcement learning based on MDP homomorphisms \citep{ravindran2004algebraic}, including \citet{van2020mdp} that integrate symmetry through an equivariant policy network. Furthermore, \citet{mondal_group_2020} previously applied a similar idea without using MDP homomorphisms. \citet{park_learning_2022} learn equivariant transition models, but do not consider planning, and \citet{zhao_toward_2022} focuses on permutation symmetry in object-oriented transition models. Recent research by \citep{zhao_integrating_2022,zhao_scaling_2023} on 2D discrete symmetry on 2D grids has used a value-based planning approach.

\subsection{Additional Related Work}

\paragraph{Geometric deep learning and equivariant networks.}

Geometric deep learning is a field that examines how to maintain geometric properties, such as symmetry and curvature, in data analysis \citep{bronstein_geometric_2021}. To preserve symmetries in data, researchers have developed equivariant neural networks. For instance, \citet{cohen_group_2016} introduced G-CNNs, followed by Steerable CNNs \citep{cohen_steerable_2016}, which generalize scalar feature fields to vector fields and the induced representations. Moreover, \citet{kondor_generalization_2018,cohen_general_2020} have studied the theory on equivariant maps and convolutions for scalar fields through trivial representations and vector fields through induced representations, respectively. Furthermore, \citet{weiler_general_2021} propose $E(2)$-CNN, a method for solving kernel constraints for $E(2)$ and its subgroups, by decomposing into irreducible representations. Researchers have also explored how to use steerable features to maintain symmetries in deep learning models. For example, \citet{brandstetter_geometric_2022} developed steerable message-passing GNNs that use equivariant steerable features, while \citet{satorras_en_2021} use only invariant scalar features to build $E(n)$-equivariant graph networks. The idea of steerable features is further developed in \citet{brandstetter_geometric_2022}, who propose steerable message passing graph networks for 3D space.

\paragraph{Learned dynamics model for interactive system.}

A framework for learning interaction dynamics between objects in a scene was proposed by \citet{battaglia_interaction_2016}. This approach is based on relational inductive biases that consider the relationships among objects. \citet{battaglia_relational_2018} later expanded on this framework by introducing relational networks that learn the dynamics between objects within a graph-based representation. Similarly, \citet{sanchez-gonzalez_graph_2018} developed a graph neural network model for physics simulation to learn dynamics in a graph-based representation. Furthermore, \citet{li_learning_2019} introduced a particle-based dynamics network that focuses on the physical interactions between particles in a simulation. This approach enables the generation of realistic animations and predictions of future states.

\subsection{Limitations and Future Work}

Although Euclidean symmetry group is infinite and seems huge, it does not guarantee significant performance gain in all cases.
Our theory helps us understand when such Euclidean symmetry may not be very beneficial
The key issue is that when a robot has kinematic constraints, Euclidean symmetry does not change those features, which means that equivariant constraints cannot share parameters and reduce dimensions.
We empirically show this on using local vs. global reference frame in the additional experiment in Sec~\ref{sec:add_results}.
For further work, one possibility is to explicit consider constraints while keep using global positions.

\subsection{Background for Representation Theory and $G$-steerable Kernels}

We establish some notation and review some elements of group theory and representation theory. For a comprehensive review of group theory and representation theory, please see \citep{Serre_2005}. The identity element of any group $G$ will be denoted as $e$. We will always work over the field $\mathbb{R}$ unless otherwise specified.

\subsection{Group Definition}
A group is a non-empty set equipped with an associative binary operation $\cdot: G \times G \rightarrow G$ where $\cdot$ satisfies
\begin{align*}
 & \text{Existence of identity: } \exists e \in G, \text{ s.t. } \forall g\in G, \enspace e \cdot g = g \cdot e = g \\
& \text{Existence of inverse: } \forall g\in G, \exists g^{-1} \in G \text{ s.t. } g \cdot g^{-1} = g^{-1} \cdot g = e
\end{align*}

For a complete reference on group theory, please see \cite{Zee_2016}.

\subsubsection{Group Representations}
A group is an abstract object. Oftentimes, when working with groups, we are most interested in group \emph{representations}.
Let $V$ be a vector space over $\mathbb{C}$. A \emph{representation} $(\rho , V)$ of $G$ is a map $\rho : G \rightarrow \Hom[V,V]$ such that 
\begin{align*}
\forall g , g' \in G, \enspace \forall v\in V, \quad   \rho( g \cdot g' )v =  \rho( g ) \cdot \rho(  g' )v
\end{align*}
Concisely, a group representation is a embedding of a group into a set of matrices. The matrix embedding must obey the multiplication rule of the group. Over $\mathbb{R}$ and $\mathbb{C}$ all representations break down into irreducible representations \cite{Serre_2005}. We will denote the set of irreducible representations of a group $G$ and $\hat{G}$.

\subsubsection{Group Actions}\label{Section:Group Actions}

Let $\Omega$ be a set. A group action $\Phi$ of $G$ on $\Omega$ is a map $\Phi : G \times \Omega \rightarrow \Omega$ which satisfies 
\begin{align*}
    &\text{Identity: } \forall \omega \in \Omega, \quad \Phi(e , \omega )  =  \omega \\
    &\nonumber \text{Compositionality: }\forall g_{1},g_{2} \in G,\enspace \forall \omega \in \Omega, \quad  \Phi( g_{1}g_{2} , \omega  ) = \Phi(g_{1} , \Phi(g_{2}, \omega ))
\end{align*}
We will often suppress the $\Phi$ function and write $\Phi( g , \omega ) = g \cdot \omega$.

\begin{center}\label{Diagram:G-Equivarient_Map}
    \begin{tikzcd}\centering
        &\Omega \arrow{d}{\Phi(g,\cdot)} \arrow{r}{ \Psi } & \Omega' \arrow{d}{ \Phi'(g, \cdot ) }  \\
        & \Omega \arrow{r}{\Psi }  & \Omega'
    \end{tikzcd}
    \captionof{figure}{Commutative Diagram For $G$-equivariant function: Let $\Phi(g, \cdot ): G \times \Omega \rightarrow \Omega$ denote the action of $G$ on $\Omega$. Let $\Phi'(g, \cdot ): G \times \Omega' \rightarrow \Omega'$ denote the action of $G$ on $\Omega'$ The map $\Psi: \Omega \rightarrow \Omega'$ is $G$-equivariant if and only if the following diagram is commutative for all $g\in G$.    }
\end{center}

Let $G$ have group action $\Phi$ on $\Omega$ and group action $\Phi'$ on $\Omega'$. A mapping $\Psi : \Omega \rightarrow \Omega'$ is said to be $G$-equivariant if and only if
\begin{align}\label{Equation:G_Equivariece_Def}
    \forall g \in G, \forall \omega \in \Omega, \quad	 \Psi(  \Phi( g , \omega )  ) = \Phi'( g , \Psi(\omega) )
\end{align}
Diagrammatically, $\Psi$ is $G$-equivariant if and only if the diagram \ref{Diagram:G-Equivarient_Map} is commutative.

\paragraph{$G$-Intertwiners}

Let $(\rho , V)$ and $(\sigma , W)$ be two $G$-representations. The set of all $G$-equivariant linear maps between $(\rho , V)$ and $(\sigma , W)$ will be denoted as
\begin{align*}
    \Hom_{G}[ (\rho , V) , (\sigma , W) ] = \{ \Phi \enspace| \enspace \Phi: V \rightarrow W, \enspace \forall g\in G, \enspace \Phi ( \rho(g) v ) = \sigma(g) \Phi(v) \}
\end{align*}
$\Hom_{G}$ is a vector space over $\mathbb{C}$. When the linear maps are restricted to be real, $\Hom_{G}$ forms a vector space over $\mathbb{R}$. A linear map $\Phi \in \Hom_{G}[ (\rho , V) , (\sigma , W) ] $ is said to \emph{intertwine} the representations $(\rho , V)$ and $(\sigma , W)$. An intertwiner $\Phi$ is a map that makes the diagram \ref{Diagram:G-Equivarient_Map} commutative.

\begin{center}\label{Diagram:G-Intertwiner}
    \begin{tikzcd}\centering
        & (\rho , V) \arrow{d}{ \rho(g) } \arrow{r}{ \Phi } & ( \sigma , W) \arrow{d}{ \sigma(g)  }  \\
        & (\rho , V) \arrow{r}{ \Phi }  & ( \sigma , W)
    \end{tikzcd}
    \captionof{figure}{Commutative Diagram For $G$-intertwiner. The map $\Psi: \in \Hom_{G}[ (\rho , V) , (\sigma , W) ] $ if and only if the following diagram is commutative for all $g\in G$.    }
\end{center}

Computing a basis for the vector space $\Hom_{G}[ (\rho , V) , (\sigma , W) ]$ is an important procedure in the theory of steerable kernels \citep{Cohen_2016}.

\subsubsection{Clebsch-Gordon Coefficients}

Let $G$ be a compact group. Let $\hat{G}$ be the irreducible representations of $G$ Let $(\rho , V)$ and  $(\sigma , W)$ be irreducible representations of $G$. The tensor product representation will not in general be irreducible and 
\begin{align*}
(\rho , V) \otimes (\sigma , W) = \bigoplus_{ \tau \in \hat{G} } c^{\tau}_{\sigma,\rho} (\tau , V_{\tau} )
\end{align*}
where $c^{\tau}_{\sigma,\rho}$ are the Clebsch-Gordon multiplicities which count the number of copies of the irreducible $(\tau , V_{\tau} )$ in the tensor product representation $(\rho , V) \otimes (\sigma , W)$. Clebsch-Gordon Coefficients $C^{\tau}_{\rho_{1}\rho_{2}}$ are the coefficients of the representation $(\tau , V_{\tau})$ in the tensor product basis. Specifically, let
\begin{align*}
    | \tau i_{\tau} \rangle = \sum_{j_{1}=1}^{d_{1}} \sum_{j_{2}=1}^{d_{2}} \underbrace{ \langle \rho_{1} j_{1} ,  \rho_{2} j_{2} | \tau i_{\tau} \rangle }_{ ( C^{\tau}_{\rho_{1}\rho_{2}} )_{i_{\tau} , j_{1}j_{2} } } | \rho_{1} j_{1},  \rho_{2} j_{2}  \rangle 
\end{align*}
Clebsch-Gordon coefficients are an integral part of the general solution to steerable kernel constraint \cite{Lang_2020}.

\subsubsection{Characterization of Steerable Kernels on Homogeneous Spaces }
    We briefly summarize the results of \citep{Lang_2020}. Let $X$ be a homogeneous space of a compact group $G$. Let $(\sigma , V_{\sigma} ) \in \hat{G}$ and $(\rho , V_{\rho} ) \in \hat{G}$ be two $G$-irreducibles. Consider the kernel constraint
    \begin{align*}
        K(g \cdot x) = \sigma(g) K(x) \rho(g^{-1})
    \end{align*}
    where $K : X \rightarrow \text{Hom}[V_{\rho},V_{\sigma}]$. Then, there exists a set of generalized spherical harmonics $Y^{i}_{\rho , k} : X \rightarrow \mathbb{R} $ where $\rho \in \hat{G}$ is a $G$-irreducible and the index $i\in\{1,2,...,d_{\rho}\}$ and $k \in \{1,2,...,m_{\rho}\}$ where $m_{\rho} \leq d_{\rho}$ is called the muplicity which satisfy the relation
    \begin{align*}
    \forall g \in G, \quad Y^{i}_{\rho , k}( g^{-1} \cdot x ) =  \sum_{i'=1}^{d_{\rho}} \rho_{ii'}(g) Y^{i'}_{\rho , k}(  x )
    \end{align*}
    The set of $Y^{i}_{\rho , k}$ form a basis for all square integrable functions on $X$.
    
    Let us define 
	\begin{align*}
		K^{\tau ks }_{\sigma \rho}(x)  =  \sum_{i_{\tau}=1}^{d_{\tau}} \sum_{j_{\sigma}=1}^{d_{\sigma}} \sum_{i_{\rho} = 1}^{d_{\rho}} |\sigma j_{\sigma} \rangle  \underbrace{ \langle s , \sigma j_{\sigma}| \tau i_{\tau}, \rho i_{\rho} \rangle }_{ \text{Clebsch-Gordon} }  \underbrace{ Y^{ i_{\rho} }_{\rho , k}(x) }_{ \text{harmonics} } \langle \rho i_{\rho} |
	\end{align*}
	Then, using the main result of \citep{Lang_2020}, the matrices $K^{\tau ks }_{\sigma \rho}(x)$ form a basis for the space of $G$-steerable kernels with input representation $\rho$ and output representation $\sigma$. Then, the kernel $K$ can be written in the form
	\begin{align*}
		K_{\sigma \rho}(x) = \sum_{\tau \in \hat{G} } \sum_{k=1}^{m_{\rho}}  \sum_{s=1}^{m_{\sigma}}  c^{\tau ks} K^{\tau ks }_{\sigma \rho}(x) 
	\end{align*}
	where $c_{\tau k s} \in \text{Hom}_{G}[ ( \sigma , V_{\sigma}) , (\sigma, V_{\sigma} ) ]$ is a $(\sigma, V_{\sigma})$-endomorphism. The total number of free parameters in $K_{\sigma \rho}$ is
	\begin{align*}
	 \dim K_{\sigma \rho} = m_{\rho}  m_{\sigma} 	\sum_{\tau \in \hat{G} } C^{\sigma}_{\tau \rho}  \times \dim \text{Hom}_{G}[ ( \sigma , V_{\sigma}) , (\sigma, V_{\sigma} ) ] \leq 4 m_{\rho} m_{\sigma} \sum_{\tau \in \hat{G} } c^{\sigma}_{\tau \rho}
	\end{align*}
    which depends on both multiplicity $m_{\tau}$ of the homogeneous space $X$ and the Clebsch-Gordon Coefficients $c^{\sigma}_{\tau \rho}$ of the group $G$.

\section{Theory and Proofs}
\label{sec:theory-appendix}

The section is organized as follows.
We first give the proofs to Theorem 1 and 2.
Then, we discuss how we linearize dynamics of a Geometric MDP and $G$-steerable kernels in detail.
The goal of the theory is to show that, in linearized case, Euclidean symmetry can provably \textit{reduce} number of free parameters and the dimensions of the solution space.
Under RL setup with \textit{unknown} dynamics (and cost) function, the Euclidean equivariance constraints then potentially bring significant benefit because of \textit{less parameters}.

\subsection{Theorem 1 and 2: Equivariance in Geometric MDPs}

\textbf{\textit{Theorem 1}}
\textit{The Bellman operator of a GMDP is equivariant under Euclidean group $\mathrm{E}(d)$.}

\textit{Proof.}
The Bellman (optimality) operator is defined as
\begin{equation}
    \gT [V](\vs)  := \max_{\va} R(\vs, \va) + \int d\vs' P(\vs' \mid \vs, \va) V(\vs'),
\end{equation}
where the input and output of the Bellman operator are both value function $V: \gS \to \sR$.
The theorem directly generalizes to $Q$-value function.

Under group transformation $g$, a feature map (field) $f: X \to \sR^{c_\text{out}}$ is transformed as:
\begin{equation}
	\left[L_g f\right](x)=\left[f \circ g^{-1}\right](x)= \rho_\text{out}(g) \cdot f\left(g^{-1} x\right),
\end{equation}
where $\rho_\text{out}$ is the $G$-representation associated with output $\sR^{c_\text{out}}$.
For the \textit{scalar} value map, $\rho_\text{out}$ is identity, or trivial representation.

For any group element $g \in \mathrm{E}(d) = \sR^d \rtimes \mathrm{O}(d)$, we transform the Bellman (optimality) operator step-by-step and show that it is equivariant under $\mathrm{E}(d)$:
\begin{align}
L_g \left[ \gT [V] \right](\vs) 
& \stackrel{\text{(1)}}{=} \gT [V](g^{-1}\vs) \\
& \stackrel{\text{(2)}}{=} \max_{\va} R(g^{-1}\vs, \va) + \int d\vs' \cdot P(\vs' \mid g^{-1} \vs, \va) V(\vs') \\
& \stackrel{\text{(3)}}{=} \max_{\bar \va} R(g^{-1}\vs, g^{-1} \bar \va) + \int d(g^{-1} \bar{\vs}) \cdot P(g^{-1} \bar{\vs} \mid g^{-1} \vs, g^{-1}\va) V(g^{-1}  \bar{\vs}) \\
& \stackrel{\text{(4)}}{=} \max_{\bar \va} R(\vs, \bar \va) + \int d(g^{-1} \bar{\vs}) \cdot P(\bar{\vs} \mid \vs, \va) V(g^{-1} \bar{\vs}) \\
& \stackrel{\text{(5)}}{=} \max_{\bar \va} R(\vs, \bar \va) + \int d\bar{\vs} \cdot P(\bar{\vs} \mid \vs, \va) V(g^{-1} \bar{\vs}) \\
& \stackrel{\text{(6)}}{=} \gT [ L_g [V]](\vs) 
\end{align}

For each step:
\begin{itemize}
\item (1) By definition of the (left) group action on the feature map $V: \gS \to \sR$, such that $g\cdot V(\vs) = \rho_0(g) V(g^{-1} \vs) = V(g^{-1} \vs)$. Because $V$ is a scalar feature map, the output transforms under trivial representation $\rho_0(g) = \mathrm{Id}$.
\item (2) Substitute in the definition of Bellman operator.
\item (3) Substitute $\va = g^{-1} (g \va) = g^{-1} \bar \va$. Also, substitute $g^{-1} \bar \vs = \vs'$.
\item (4) Use the symmetry properties of Geometric MDP: $P(\vs' \mid \vs, \va) = P(g \cdot \vs \mid g \cdot \vs, g \cdot \va)$ and $R (\vs, \va) = R(g \cdot \vs, g \cdot \va)$.
\item (5) Because $g \in \mathrm{E}(d)$ is isometric transformations (translations $\sR^d$, rotations and reflections $\mathrm{O}(d)$) and the state space carries group action, the measure $ds$ is a $G$-invariant measure $d(gs) = ds$. Thus, $d \bar{\vs} = d(g^{-1} \bar{\vs})$. 
\item (6) By the definition of the group action on $V$.
\end{itemize}

The proof requires the MDP to be a Geometric MDP with Euclidean symmetry and the state space carries a group action of Euclidean group.
Therefore, the Bellman operator of a Geometric MDP is $\mathrm{E}(d)$-equivariant.
Additionally, we can also parameterize the dynamics and reward functions with neural networks, and the learned Bellman operator is also equivariant.

The proof is analogous to the case in \citep{zhao_integrating_2022}, where the symmetry group is $p4m=\sZ^2 \rtimes D_4$, which is a discretized subgroup of $\mathrm{E}(2)$.
A similar statement can also be found in symmetric MDP \citep{zinkevich_symmetry_2001}, MDP homomorphism induced from symmetry group \citep{ravindran2004algebraic}, and later work on symmetry in deep RL \citep{van_der_pol_mdp_2020,wang_mathrmso2-equivariant_2021}.

\textbf{\textit{Theorem 2}}
\textit{For a GMDP, value iteration is an $\mathrm{E}(d)$-equivariant geometric message passing.}

\begin{figure*}[t]
\centering
\subfigure{

\includegraphics[width=.5\linewidth]{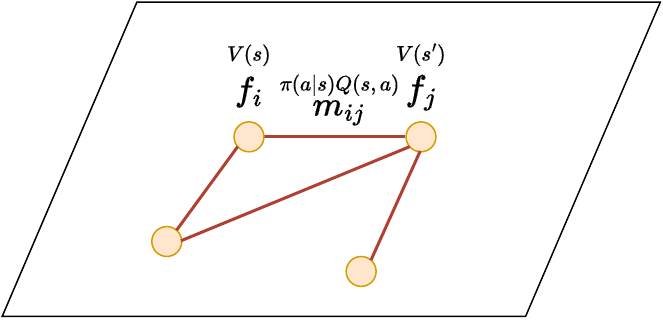}
}
\centering
\caption{
\small
Demonstrate the idea of implementing value iteration with geometric message passing.
\todoilzlf{add caption a bit more}
}
\label{fig:demo-vi-mp}
\end{figure*}

\textit{Proof.}
We prove by constructing value iteration with 
For a more rigorous account on the relationship between dynamic programming (DP) and message passing on \textit{non-geometric} MDPs, see \citep{dudzik_graph_2022}.

Notice that they satisfy the following equivariance conditions:
\begin{align}
P_\theta &: \gS \times \gA \times \gS \to \sR^+  :& P_\theta(\vs_{t+1} \mid \vs_{t}, \va_{t}) &=  P_\theta(\rho_\gS(g) \cdot \vs_{t+1} \mid \rho_\gS(g) \cdot \vs_{t}, \rho_\gA(g) \cdot \va_{t}) \\
R_\theta &: \gS \times \gA \to \sR :& {R}_\theta(\vs_{t}, \va_{t} ) &=  {R}_\theta(\rho_\gS(g) \cdot \vs_{t}, \rho_\gA(g) \cdot \va_{t}) \\
Q_\theta &: \gS \times \gA \to \sR :& {Q}_\theta(\vs_{t}, \va_{t} ) &=  {Q}_\theta(\rho_\gS(g) \cdot \vs_{t}, \rho_\gA(g) \cdot \va_{t}) \\
V_\theta &: \gS \to \sR :& {V}_\theta(\vs_{t}) &=  {V}_\theta(\rho_\gS(g) \cdot \vs_{t}) \\
\end{align}

We construct geometric message passing such that it uses \textit{scalar} messages and features and resembles value iteration.
The idea is visualized in Fig~\ref{fig:demo-vi-mp}.

Then, we can use geometric message passing network to construct value iteration, which is to iteratively apply Bellman operators.
We adopt the definition of geometric message passing based on \citep{brandstetter_geometric_2021} as follows.
\begin{align}
\tilde{\mathbf{m}}_{i j} & =\phi_m\left(\tilde{\mathbf{f}}_i, \tilde{\mathbf{f}}_j, \tilde{\mathbf{a}}_{i j}\right) \\
\tilde{\mathbf{f}}_i^{\prime} & =\phi_f\left(\tilde{\mathbf{f}}_i, \sum_{j \in \mathcal{N}(i)} \tilde{\mathbf{m}}_{i j}, \tilde{\mathbf{a}}_i\right) .
\end{align}
The tilde means they are steerable under $G$ transformations.

We want to construct value iteration:
\begin{align}
	Q(s,a) &= R(s,a) + \gamma \sum_{s'} P(s'|s, a) V(s') \\
	V'(s) &= \sum_a \pi(a|s) Q(s,a)
\end{align}

To construct a \textit{geometric graph}, we let vertices $\gV$ be states $\vs$ and edges $\gE$ be state-action transition $(\vs, \va)$ labelled by $\va$.
For the geometric features on the graph, there are node features and edge features.
Node features include maps/functions on the state space: $\gS \to \sR^D$, and edge features include functions on the state-action space $\gS \times \gA \to \sR^D$.

For example, state value function $V: \gS \to \sR$ is (scalar) node feature, and $Q$-value function $Q_\theta: \gS \times \gA \to \sR$ and reward function $R_\theta : \gS \times \gA \to \sR$ are edge features.
The message $\tilde{\mathbf{m}}_{i j}$ is thus a scalar for every edge: $\tilde{\mathbf{m}}_{i j} = \pi(a|s) Q(s,a)$, and $\tilde{\mathbf{f}}_i^{\prime}$ is updated value function $\tilde{\mathbf{f}}_i^{\prime} = V'(s)$.
It is possible to extend value iteration to vector form as in Symmetric Value Iteration Network and Theorem 5.2 in \citep{zhao_integrating_2022}, while we leave it for future work.

\subsection{Linear-Quadratic Control: Linearizing Geometric MDPs}
\label{subsec:lqr-linearization}

Linear-Quadratic Regulator (LQR) is one of the most frequently used methods in optimal control \cite{Tedrake_2023}. 
LQR is a computationally efficient method for solving problems with linear dynamics and quadratic costs.
LQR has various noise robustness and optimally guarantees.
Even if the dynamical system is nonlinear, linear-quadratic control methods have been used after iteratively linearizing the dynamics and quadratizing the cost.

Many of the problems where LQR is applied have symmetries. Recently has the control community began to study how symmetry can be used to increase the performance of classical control algorithms \cite{teng_error-state_2023,Teng_2022,Ghaffari_2022}. \cite{Hampsey_2022,Hampsey_2022_II,Cohen_2020_I} specifically consider LQR on homogeneous spaces but do not establish the connection to steerable kernels. 

\edit{We show how Euclidean symmetry inherently simplifies the linearized problem.}
We assume the dynamics and cost (reward) are \textbf{unknown} and need to be \textit{learned}, thus equivariance constraints come in and \textit{reduce the number of free parameters}.

\subsection{Theorem 3: Equivariance of Linearized Dynamics}

We show the derivation of steerable kernel constraint in this subsection and further discuss the characteristics of steerable kernels in the next subsection.

\textbf{\textit{Theorem 3:}}
\textit{If a Geometric MDP has an infinitesimal $G$-action on the state-action space $\gS \times \gA$, the linearized dynamics is also $G$-equivariant: the matrix-value functions $A : \mathcal{S} \times \mathcal{A} \rightarrow \mathbb{R}^{d_{\mathcal{S} } \times d_{ \mathcal{S} }}$ and $B : \mathcal{S} \times \mathcal{A} \rightarrow \mathbb{R}^{d_{ \mathcal{S} }\times d_{ \mathcal{A} }}$ satisfy $G$-steerable kernel constraints.}

Oftentimes, the problem of interest has continuous symmetry $G$ that acts on the space $\mathcal{S} \times \mathcal{A}$. We will assume that the Lie group action $G \times (\mathcal{S} \times \mathcal{A}) \rightarrow \mathcal{S} \times \mathcal{A} $ is continuous.

Under infinitesimal symmetry transformation $g \approx 1_{G} \in G$, let us suppose that the state and action space transform as
\begin{align*}
\vs \rightarrow \rho_{\gS}(g) \cdot \vs , \quad \va \rightarrow \rho_{\gA}(g) \cdot \va
\end{align*}
where $\rho_{\gS}$ and $\rho_{\gA}$ are representations of the group $G$. Let $A$ and $B$ be the lineariziations of the dynamics $f$ at the point $p = ( \vs , \va) $, with
 \begin{align*}
     A(p) = \frac{\partial f }{\partial s} \rvert_{p}, \quad B(p) = \frac{\partial f }{\partial a} \rvert_{p}
 \end{align*}

Under infinitesimal symmetry transformation $g \approx 1_{G} \in G$, the dynamics must satisfy,
\begin{align}\label{eq:appendix-nonlinear-dyn}
\rho_{\gS}(g) \cdot f( \vs ,  \va ) = f( \rho_{\gS}(g)   \cdot \vs , \rho_{\gA}(g)   \cdot \va )
\end{align}

Because we assume the state and action space has \textit{continuous} group action, we can apply Taylor expansion for continuous group actions and only keep the first order term. Let $g = 1_{G} + \delta g + \mathcal{ O }( \delta g^{2} )$ be the expansion around the identity element, then 
\begin{align*}
& \rho_{\gS}(g) = \rho_{\gS}(1_{G}) + \rho_{\gS}( \delta g ) + \mathcal{O}( \delta g^{2}) = \mathbf{1}_{ d_{\gS} } + \rho_{\gS}( \delta g ) + \mathcal{O}( \delta g^{2} ) \\
& \rho_{\gA}(g) = \rho_{\gA}(1_{G}) + \rho_{\gA}( \delta g ) + \mathcal{O}( \delta g^{2}) = \mathbf{1}_{d_{\gA}}  + \rho_{\gA}( \delta g ) + \mathcal{O}( \delta g^{2})
\end{align*}

inserting into \ref{eq:appendix-nonlinear-dyn}, and collecting terms of order $\mathcal{O}(\delta g)$, we have that,
\begin{align*}
\rho_{\gS}(\delta g) \frac{\partial f }{\partial s} \rvert_{\delta p} = \frac{\partial f }{\partial s} \rvert_{ \delta g \cdot p} \rho_{\gS}(\delta g), \quad \rho_{\gS}(\delta g) \frac{\partial f }{\partial a} \rvert_{p} = \frac{\partial f }{\partial a} \rvert_{ \delta g \cdot p} \rho_{\gS}(\delta g)
\end{align*}
 
In general, we solve the non-linear problem by linearizing and iterating. Let us consider the linearized problem at point $p = ( \vs_{0} , \va_{0} )$. Assuming that the state and control do not change too drastically over a short period of time, we can approximate the true dynamics as
	\begin{align*}
		\vs_{t+1} = A( p ) \cdot  \vs_{t} + B( p ) \cdot \va_{t}
	\end{align*}
    where $A : \mathcal{S} \times \mathcal{A} \rightarrow \mathbb{R}^{d_{\mathcal{S} } \times d_{ \mathcal{S} }}$ and $B : \mathcal{S} \times \mathcal{A} \rightarrow \mathbb{R}^{d_{ \mathcal{S} }\times d_{ \mathcal{A} }}$.

    Now, linearizing the dynamics $f$ and using the symmetry constraint.
The matrix valued functions $A(p)$ and $B(p)$ must satisfy the constraints
	\begin{align}\label{Equation: A and B kernel}
		\forall g\in G, \quad	A( g \cdot p ) = \rho_{\gS}(g) A( p ) \rho_{\gS}(g^{-1}) , \quad B( g \cdot p ) = \rho_{\gS}(g) B( p ) \rho_{\gA}(g^{-1}) 
	\end{align}
	so that $A$ is a $G$-steerable kernel with input representation $\rho_{\gS}$ and output representation $\rho_{\gS}$ and $B$ is a $G$-steerable kernel with input representation $\rho_{\gA}$ and output representation $\rho_{\gS}$ \cite{Cohen_2016}. The kernel constraints \ref{Equation: A and B kernel} relate $A( p )$ and $B( p )$ at different points. 
 We use this constraint to understand why Euclidean symmetry is beneficial for decision-making: it \textit{reduces} the number of free parameters and the dimensions of the solution space.

\subsection{Characteristics of $G$-Steerable Kernels}

Consider the $G$-steerable kernel constraints,
	\begin{align}\label{Equation: A and B constraints }
		\forall g\in G, \quad	A( g \cdot p) = \rho_{\gS}(g) A(p) \rho_{\gS}(g^{-1}) , \quad B( g \cdot p ) = \rho_{\gS}(g) B( p) \rho_{\gA}(g^{-1}) 
	\end{align}
	To begin, note that this equation only relates $A$ and $B$ on points that can be related by $G$-transformation. As first observed in \cite{Weiler_2018}, we can solve the constraints \ref{Equation: A and B constraints } over each orbit separately. Let us define the $G$-orbits of $ \mathcal{S} \times \mathcal{A} $ as
    \begin{align}
    O(x) = \{ \enspace y \enspace | \enspace \exists g \in G,\enspace y = g \cdot x \enspace \}
    \end{align}
    where $x\in \mathcal{S} \times \mathcal{A} $. Every $G$-orbit is a homogeneous space of $G$ \cite{Serre_2005}. The set of $G$-orbits form a partition of the space $ \mathcal{S} \times \mathcal{A}$. Let us define the equivalence relation $\sim$ as
    \begin{align*}
	   x \sim y \implies \exists g\in G, \text{ such that } x = g \cdot y
	\end{align*}
    so that $x \sim y$ only if $x$ and $y$ are related by symmetry transformation. We then define the quotient space 
    \begin{align*}
        \base = ( \mathcal{S} \times \mathcal{A} ) / \sim
    \end{align*}
    where the space $\base$ consists of the space of all $G$-orbit representatives. Using a standard result in topology \cite{Husemöller_2013}, there is then a canonical continuous projection map $\Pi : \mathcal{S} \times \mathcal{A} \rightarrow \base $ which projects each point in $\mathcal{S} \times \mathcal{A} $ to a canonically chosen orbit representative. Note that points related by $G$-action have the same projection and
    \begin{align*}
     \forall g \in G, \quad \Pi( g \cdot x) = \Pi(x)    
    \end{align*}
    holds for all $x\in \mathcal{S} \times \mathcal{A}$. We can decompose every point in the space $\mathcal{S} \times \mathcal{A}$ into an element of $\base$ and a element of a homogeneous space of $G$. Let $p \in \mathcal{S} \times \mathcal{A}$, we can always write
    \begin{align*}
        p = x_{p} \times p_{\downarrow}
    \end{align*}
    where $ p_{\downarrow} = \Pi(p) \in \base$ is the orbit representative of $p$ and $x_{p}$ is an element of $O( p_{\downarrow} )$ which is a homogeneous space of $G$. Now, using this decomposition, we may write the constraints \ref{Equation: A and B kernel} as
	\begin{align*}
	\forall g \in G, \quad	A(g \cdot x , p_{\downarrow} ) = \rho_{\gS}(g) A( x , p_{\downarrow} )  \rho_{\gS}(g^{-1}), \quad B(g \cdot x , p_{\downarrow} ) = \rho_{\gS}(g) B( x , p_{\downarrow} )  \rho_{\gA}(g^{-1}),
	\end{align*}
    A complete solution to this constraint for any compact group $G$ was given in \cite{Lang_2020}. Following \cite{Weiler_2018}, we can simplify these kernel constraints based on the decomposition of $\rho_{\gS}$ and $\rho_{\gA}$ into irreducibles. Let $\hat{G}$ denote a representative set of $G$-irreducibles. Let us suppose that $\rho_{\gS}$ and $\rho_{\gA}$ decompose into irreducibles of $G$ as
	\begin{align*}
		(\rho_{\gS} , V_{\gS} ) = \bigoplus_{ \sigma \in \hat{G} } n_{\sigma} (\sigma , V_{\sigma}) , \quad (\rho_{\gA},V_{\gA} ) = \bigoplus_{\sigma \in  \hat{G} } q_{\sigma}(\sigma,V_{\sigma}) 
	\end{align*}
	where $(\sigma , V_{\sigma}) $ are the $G$-irreducibles and $n_{\sigma}$ and $q_{\sigma}$ count the multiplicity of the $(\sigma , V_{\sigma}) $ irreducible in $\rho_{\gS}$ and $\rho_{\gA}$, respectively. The dimensions of each irreducible are related to the dimensions of the $\mathcal{S}$ and $\mathcal{A}$ manifolds via
	\begin{align*}
		d_{\mathcal{S}} = \sum_{ \sigma \in \hat{G} }  d_{\sigma} n_{\sigma}, \quad \quad d_{ \mathcal{A} } = \sum_{ \sigma \in \hat{G} } d_{\sigma} q_{\sigma}
	\end{align*}
    Now, by definition of an reducible representation, there exists a $d_{\mathcal{S} } \times d_{\mathcal{S} }$ unitary matrix $U$ and a $d_{\mathcal{A} } \times d_{\mathcal{A} } $ unitary matrix $V$ such that we change basis and write
	\begin{align*}
	\forall g\in G, \quad	\rho_{\gS}(g) = U \begin{bmatrix}
			n_{1} \sigma_{1}( g ) & 0 & 0 &... &0 &0 \\
			0 & 	n_{2} \sigma_{2}( g ) & 0 &...& 0 &0 \\
			0 & 	0 & n_{3} \sigma_{3}( g ) &...& 0 &0 \\
			... & 	... & ... &... & ... & ... \\
			0 & 	& 0 &... &0 & n_{|\hat{G}|} \sigma_{|\hat{G}|}( g )  \\
		\end{bmatrix} U^{\dagger} 	
	\end{align*}
	and 
	\begin{align*}
	\forall g\in G, \quad	\rho_{\gA}(g) = V \begin{bmatrix}
			q_{1} \sigma_{1}( g ) & 0 & 0 &... &0 &0 \\
			0 & 	q_{2} \sigma_{2}( g ) & 0 &...& 0 &0 \\
			0 & 	0 & q_{3} \sigma_{3}( g ) &...& 0 &0 \\
			... & 	... & ... &... & ... & ... \\
			0 & 	& 0 &... &0 & q_{|\hat{G}|} \sigma_{|\hat{G}|}( g )  \\
		\end{bmatrix} V^{\dagger} 
	\end{align*}
	where the notation $b_{i}\sigma_{i}( g )$ denotes a block matrix with $b_{i}$ copies of the irreducible $( \sigma_{i} , V_{i} ) \in \hat{G}$ on the diagonals,
	\begin{align*}
		\forall g\in G, \quad	b_{i}\sigma_{i}(g) = \underbrace{ \begin{bmatrix}
			\sigma_{i}( g ) & 0 & 0 &... &0 &0 \\
			0 &  \sigma_{i}( g ) & 0 &...& 0 &0 \\
			0 &  0 & \sigma_{i}( g ) &...& 0 &0 \\
			... & 	... & ... &... & ... & ... \\
			0 & 	& 0 &... &0 &  \sigma_{i}( g )  \\
		\end{bmatrix} }_{ b_{i}\text{ copies} }
	\end{align*}

    We can write down the solution explicitly in this basis,
	\begin{align*}
	&	A(x, p_{\downarrow})  = U \begin{bmatrix}
			n_{1} n_{1} K_{11}^{A}(x, p_{\downarrow}) & n_{1} n_{2} K_{12}^{A}(x, p_{\downarrow}) & n_{1} n_{3} K^{A}_{13}(x, p_{\downarrow}) &...& n_{1} n_{|\hat{G}|} K^{A}_{1|\hat{G}|}(x, p_{\downarrow}) \\
			n_{2} n_{1} K^{A}_{21}(x, p_{\downarrow}) & n_{2} n_{2} K^{A}_{22}(x, p_{\downarrow}) & n_{2} n_{3} K^{A}_{23}(x, p_{\downarrow}) &...& n_{2} n_{|\hat{G}|} K^{A}_{2|\hat{G}|}(x, p_{\downarrow}) \\
			... & ... & ... &...& ... \\
			n_{|\hat{G}|} n_{1} K^{A}_{|\hat{G}|1}(x, p_{\downarrow}) & n_{|\hat{G}|} n_{2} K^{A}_{|\hat{G}|2}(x, p_{\downarrow}) & n_{|\hat{G}|} n_{3} K^{A}_{|\hat{G}|3}(x, p_{\downarrow}) &...& n_{|\hat{G}|} n_{|\hat{G}|} K^{A}_{|\hat{G}||\hat{G}|}(x, p_{\downarrow}) \\
		\end{bmatrix} U^{\dagger} \\
		& \text{and}\\
		&	B(x, p_{\downarrow})  = U \begin{bmatrix}
			n_{1} q_{1} K_{11}^{B}(x, p_{\downarrow}) & n_{1} q_{2} K_{12}^{B}(x, p_{\downarrow}) & n_{1} q_{3} K^{B}_{13}(x, p_{\downarrow}) &...& n_{1} q_{|\hat{G}|} K^{B}_{1|\hat{G}|}(x, p_{\downarrow}) \\
			n_{2} q_{1} K^{B}_{21}(x, p_{\downarrow}) & n_{2} q_{2} K^{B}_{22}(x, p_{\downarrow}) & n_{2} q_{3} K^{B}_{23}(x, p_{\downarrow}) &...& n_{2} q_{|\hat{G}|} K^{B}_{2|\hat{G}|}(x, p_{\downarrow}) \\
			... & ... & ... &...& ... \\
			n_{|\hat{G}|} q_{1} K^{B}_{|\hat{G}|1}(x, p_{\downarrow}) & n_{|\hat{G}|} q_{2} K^{B}_{|\hat{G}|2}(x, p_{\downarrow}) & q_{|\hat{G}|} n_{3} K^{B}_{|\hat{G}|3}(x, p_{\downarrow}) &...& n_{|\hat{G}|} q_{|\hat{G}|} K^{B}_{|\hat{G}||\hat{G}|}(x, p_{\downarrow}) \\
		\end{bmatrix} V^{\dagger}
	\end{align*}
	where $K_{ij}^{A}$ are the $A$-kernels with input representation $(\sigma_{i} , V_{i})$ and output representation $(\sigma_{j} , V_{j})$ and $K_{ij}^{B}$ are the $B$-kernels with input representation $(\sigma_{i} , V_{i})$ and output representation $(\sigma_{j} , V_{j})$. The notation $b_{i}c_{j} K_{ij}^{C}(x)$ denotes $b_{i} \times c_{j}$ independent copies of the $K_{ij}^{C}$ kernel,
	\begin{align*}
	b_{i}c_{j} K_{ij}^{C}(x, p_{\downarrow})  = \kern .4em
		\underbrace{
			\kern-.4em
			\left.
			\left[
			\begin{matrix}
				K_{ij}^{C}(x, p_{\downarrow}) &  K_{ij}^{C}(x, p_{\downarrow})  & K_{ij}^{C}(x, p_{\downarrow}) & ... & K_{ij}^{C}(x, p_{\downarrow}) \\
				 K_{ij}^{C}(x, p_{\downarrow}) &  K_{ij}^{C}(x, p_{\downarrow}) & K_{ij}^{C}(x, p_{\downarrow}) & ... & K_{ij}^{C}(x, p_{\downarrow}) \\
				... & ...   & ... &  ... & ...  \\
				K_{ij}^{C}(x, p_{\downarrow}) &  K_{ij}^{C}(x, p_{\downarrow}) & K_{ij}^{C}(x, p_{\downarrow}) & ... & K_{ij}^{C}(x, p_{\downarrow}) \\
			\end{matrix}
			\right]    
			\right\} b_{i} \text{ copies}
			\kern -5.0em
		}_{ c_{j} \text{ copies}}
		\kern 4.5em
	\end{align*}
    Using Bra-Ket notation, let us define 
	\begin{align*}
		K^{\tau ks }_{\sigma \rho}(x)  =  \sum_{i_{\tau}=1}^{d_{\tau}} \sum_{j_{\sigma}=1}^{d_{\sigma}} \sum_{i_{\rho} = 1}^{d_{\rho}} |\sigma j_{\sigma} \rangle  \underbrace{ \langle s , \sigma j_{\sigma}| \tau i_{\tau}, \rho i_{\rho} \rangle }_{ \text{Clebsch-Gordon} }  \underbrace{ Y^{ i_{\rho} }_{\rho , k}(x) }_{ \text{harmonics} } \langle \rho i_{\rho} |
	\end{align*}
	Then, using the main result of \cite{Lang_2020}, the matrices $K^{\tau ks }_{\sigma \rho}(x)$ form a basis for the space of $G$-steerable kernels with input representation $\rho$ and output representation $\sigma$. Then, the kernels $K^{A}$ and $K^{B}$ can be written in the form
	\begin{align*}
	& K^{A}_{\sigma \rho}(x) = \sum_{\tau \in \hat{G} } \sum_{k=1}^{m_{\rho}}  \sum_{s=1}^{m_{\sigma}}  c^{A}_{\tau ks}( p_{\downarrow} ) K^{\tau ks }_{\sigma \rho}(x)  \\
    & K^{A}_{\sigma \rho}(x) = \sum_{\tau \in \hat{G} } \sum_{k=1}^{m_{\rho}}  \sum_{s=1}^{m_{\sigma}}  c^{B}_{\tau ks}( p_{\downarrow} ) K^{\tau ks }_{\sigma \rho}(x)  
	\end{align*}
	where $c^{A}_{\tau k s}( p_{\downarrow}) : \base \rightarrow \text{Hom}_{G}[ ( \sigma , V_{\sigma}) , (\sigma, V_{\sigma} ) ]$ and $c^{B}_{\tau k s}( p_{\downarrow}) : \base \rightarrow \text{Hom}_{G}[ ( \sigma , V_{\sigma}) , (\sigma, V_{\sigma} ) ]$ are maps from the quotient space $\base$ into $(\sigma, V_{\sigma})$-endomorphisms.

	Thus, when $p_{\downarrow} \in \base$ is fixed, the total number of free parameters in the $A(x,p_{\downarrow} )$ matrix is
	\begin{align*}
	\dim A(x,p_{\downarrow}) = \sum_{  \rho \sigma \in \hat{G} } n_{\rho}n_{\sigma} \sum_{\tau \in \hat{G} } m_{\sigma} m_{\rho }C^{\sigma}_{\tau \rho}  \times \dim \text{Hom}_{G}[ ( \sigma , V_{\sigma}) , (\sigma, V_{\sigma} ) ] \leq 4 \sum_{  \rho \tau \sigma \in \hat{G} } n_{\rho}n_{\sigma} C^{\sigma}_{\tau \rho} m_{\sigma} m_{\rho}
	\end{align*}
	Similarly, at fixed $p_{\downarrow} \in \base$, the total number of free parameters in $B(x,p_{\downarrow} )$
	\begin{align*}
		\dim B(x,p_{\downarrow}) = \sum_{  \rho \sigma \in \hat{G} } n_{\rho}q_{\sigma} \sum_{\tau \in \hat{G} } C^{\sigma}_{\tau \rho} m_{\sigma}m_{\rho}  \times \dim \text{Hom}_{G}[ ( \sigma , V_{\sigma}) , (\sigma, V_{\sigma} ) ] \leq 4 \sum_{  \rho \tau \sigma \in \hat{G} } n_{\rho}q_{\sigma} C^{\sigma}_{\tau \rho} m_{\sigma}m_{\rho}
	\end{align*}
    Note that for fixed $p_{\downarrow}$,
    \begin{align*}
     \dim A(x,p_{\downarrow}) \leq d_{ \mathcal{S} }^{2}, \quad  \dim B(x,p_{\downarrow}) \leq d_{ \mathcal{S} } d_{ \mathcal{A} } ,
    \end{align*}
	always hold.
    To summarize, symmetry constraints force the LQR matrices $A$ and $B$ to take the form
    \begin{align*}
    A : \base \rightarrow \mathbb{R}^{ d_{S}\times d_{S} }, \quad B : \base \rightarrow \mathbb{R}^{ d_{S}\times d_{A} }
    \end{align*}
    Furthermore, the output matrices take the form of a $G$-steerable kernel \cite{Lang_2020} and are parameterized by only a small number of parameters. This should be contrasted with the non-equivarient case where
    \begin{align*}
    A : \mathcal{S} \times \mathcal{A} \rightarrow \mathbb{R}^{ d_{S}\times d_{S} }, \quad B : \mathcal{S} \times \mathcal{A} \rightarrow \mathbb{R}^{ d_{S}\times d_{A} }
    \end{align*}
    The dimension of the space $\base$ can be significantly less than that of $\mathcal{S}\times \mathcal{A}$. Symmetry constraints thus highly restricts the allowed form of LQR.

\subsection{Theorem 4: Symmetry in Solutions of LQR}

\textbf{\textit{Theorem 4:}}
\textit{The LQR feedback matrix in $a_t^\star = - K(p) s_t$ and value matrix in $V = s_t^\top P(p) s_t$ are $G$-steerable kernels, or matrix-valued functions: $K : \mathcal{S} \times \mathcal{A} \rightarrow \mathbb{R}^{d_{\mathcal{A} } \times d_{ \mathcal{S} }}$ and $P : \mathcal{S} \times \mathcal{A} \rightarrow \mathbb{R}^{d_{ \mathcal{S} }\times d_{ \mathcal{S} }}$.}

\textit{Proof.}
The solution of LQR is derived from Bellman equation. 
The results, or optimal value function $V_t(\vs) = \vs^\top P_t \vs$ and optimal policy function (feedback control law) $\va^\star = - K_t \vs$, are given by the discrete algebraic Riccati equation (DARE).

\begin{equation}
	V_t(\vs) = \vs^\top P_t \vs, \quad \va^\star = - K_t \vs,
\end{equation}
where
\begin{align}
P_t &=Q +A^\top P_{t+1} A-A^\top P_{t+1} B\left(R+B^\top P_{t+1} B\right)^{-1} B^\top P_{t+1} A \\
K_t &= \left(R+B^\top P_{t+1} B\right)^{-1} B^\top P_{t+1} A.
\end{align}
The goal is to prove that $P_t$ and $K_t$ are $G$-steerable kernels.

If we iteratively linearize the problem, similar to the linearized dynamics, we quadratize the cost function around a point $p$.
The quadratic cost also depends on the point $p$ and is given as follows.
\begin{align}
\vs_{t+1} &= A {\color{red}(p)} \cdot  \vs_{t} + B {\color{red}(p)} \cdot \va_{t}, \quad A : \mathcal{S} \times \mathcal{A} \rightarrow \mathbb{R}^{d_{\mathcal{S} } \times d_{ \mathcal{S} }}, \quad B : \mathcal{S} \times \mathcal{A} \rightarrow \mathbb{R}^{d_{ \mathcal{S} }\times d_{ \mathcal{A} }} \\
c(s,a) &= \vs^\top Q{\color{red}(p)} \vs + \va^\top R{\color{red}(p)} \va, \quad Q : \mathcal{S} \times \mathcal{A} \rightarrow \mathbb{R}^{d_{\mathcal{S} } \times d_{ \mathcal{S} }}, \quad R : \mathcal{S} \times \mathcal{A} \rightarrow \mathbb{R}^{d_{ \mathcal{A} }\times d_{ \mathcal{A} }}
\end{align}
Analogously, $Q$ and $R$ are also $G$-steerable kernels by assuming the scalar function $c(s,a)$ is $G$-invariant:
\begin{equation}
    \forall g\in G, \quad	Q {\color{red} (g \cdot p) } = \rho_\gS(g) Q {\color{red}(p)} \rho_\gS(g^{-1}) , \quad R {\color{red} (g \cdot p) } = \rho_\gA(g) R {\color{red}(p)} \rho_\gA(g^{-1}) 
\end{equation}

We prove by induction.
For simplicity, we denote $A_p := A(p)$ and similarly for $B_p, Q_p, R_p$.
We start from $P_T = Q_p$.
By the property of $G$-steerable kernel, it is $Q_p = \rho_\gS(g) Q_{g\cdot p} \rho_\gS(g^{-1})$, thus $P_T$ is a steerable kernel.

By induction, we assume $P_{t+1}$ is steerable kernel: $P_{t+1}(p) = \rho_\gS(g) P_{t+1}({g\cdot p}) \rho_\gS(g^{-1})$.
We show how each component is transformed under group transformation.
\begin{equation}
	P_t(p)=\underbrace{Q_p}_{(1)} + \underbrace{A_p^\top P_{t+1}(p) A_p}_{(2)} - \underbrace{A_p^\top P_{t+1}(p) B_p}_{(3)}  \underbrace{\left(R_p+B_p^\top P_{t+1}(p) B_p\right)^{-1}}_{(4)} \underbrace{B_p^\top P_{t+1}(p) A_p}_{(5)}
\end{equation}

For (1), by the property of $G$-steerable kernel, it is $Q_p = \rho_\gS(g) Q_{g\cdot p} \rho_\gS(g^{-1})$.

For (2), we have
\begin{align}
A_p^\top P_{t+1}(p) A_p
&= \left(\rho_\gS(g) A_{g\cdot p} \rho_\gS\left(g^{-1}\right)\right)^{\top} \left(\rho_\gS(g) P_t({g\cdot p}\right) \rho_\gS(g^{-1})) \left(\rho_\gS(g) A_{g\cdot p} \rho_\gS\left(g^{-1}\right)\right) \\
&= \rho_\gS(g) A^\top_{g\cdot p} \rho_\gS\left(g^{-1}\right)\rho_\gS(g) P_t({g\cdot p} \rho_\gS(g^{-1})) 
\rho_\gS(g) A_{g\cdot p} \rho_\gS\left(g^{-1}\right) \\
&= \rho_\gS(g) \left( A^\top_{g\cdot p} P_t({g\cdot p})  A_{g\cdot p} \right) \rho_\gS\left(g^{-1}\right).
\end{align}

We can similarly show for the rest:
\begin{align}
\text{(3)} &= 
\rho_\gS(g)
\left( A_{g \cdot p}^\top P_{t+1}(g \cdot p) B_{g \cdot p} \right)
\rho_\gA(g^{-1}) \\
\text{(4)}^{-1} &= 
\rho_\gA(g)
\left(R_{g \cdot p}+B_{g \cdot p}^\top P_{t+1}(g \cdot p) B_{g \cdot p}\right) 
\rho_\gA(g^{-1}) \\
  \text{(5)} &= 
\rho_\gA(g)
\left(B_{g \cdot p}^\top P_{t+1}(g \cdot p) A_{g \cdot p}  \right) 
\rho_\gS(g^{-1})  
\end{align}
Thus, the multiplication (3)(4)(5) transforms under input representation $\rho_\gS(g)$ and output representation $\rho_\gS(g^{-1})$.
Analogously, all terms (1), (2), and (3)(4)(5) transforms under input representation $\rho_\gS(g)$ and output representation $\rho_\gS(g^{-1})$, same for the sum.
By moving the terms, we obtain $P_{t}(p) = \rho_\gS(g) P_t({g\cdot p}) \rho_\gS(g^{-1})$ that $P_t$ is a steerable kernel.

\note{Analogously, we can show that $K$ transforms under input representation $\rho_\gS(g^{-1})$ and output representation $\rho_\gA(g)$}.

\subsection{Illustration and Examples}

We work out a few examples of how symmetry can reduce the dimensionality of the LQR problem. We specifically consider two simple problems: a free particle moving in the plane and a free particle moving in space. The agent can control an applied force and tries to steer the particle to the origin.

\begin{table}[t]
\caption{Equivariant dimension reduction for linearized dynamics. 
\edit{This table highlights the reduced dimensions of spaces of kernels. $\gX( \mathcal{M} )$ denotes the dimension of signals living on the manifold $\mathcal{M}$. }
}
\small
\centering
\begin{tabular}{l|llllllll}
\toprule
Task                       & $\gS$                     & $\gA$                                    & $G$        & $\rho_\gS$   & $\rho_\gA$   & $ \gX( \gS \times \gA ) $  & $ \gX( \base ) $                                           \\ \midrule
Free Particle in 2D & $\rtwo \times \rtwo $     & $ \rtwo $      & $\sotwo$   & $\rho_{std}$ & $\rho_{std}$ & $ \mathbb{R}^{16} $ & $ \mathbb{R}^{+} \times \mathbb{R}^{14}  $                  \\
Reacher (in 2D) & $ S^{1} \times S^{1} \times \rtwo $     & $ \rtwo $      & $\sotwo$   & $\rho_{std} \oplus \rho_{triv}$ & $\rho_{triv}$ & $ S^{1} \times S^{1} \times \mathbb{R}^{10} $ & $ S^{1} \times \mathbb{R}^{10}   $                  \\
Single Free Particle in 3D     & $\rthree \times \rthree $ & $\rthree$ & $\sothree$ & $\rho_{std}$ & $\rho_{std}$ & $\mathbb{R}^{32}$   & $(\mathbb{R}^{+} )^{2}\times \mathbb{R}^{10} $ \\
$N$-Free Particles in 3D     & $ \mathbb{R}^{3N} \times \mathbb{R}^{3N} $ & $\mathbb{R}^{3N}$ & $\sothree$ & $\rho_{std}$ & $\rho_{std}$ & $\mathbb{R}^{32N}$   & $(\mathbb{R}^{+} )^{2}\times \mathbb{R}^{10N} $ \\ \bottomrule
\end{tabular}
\label{tab:examples-lqr-app} %
\end{table}

\paragraph{Examples.}
In Table~\ref{tab:examples-lqr-app}, we show the dimensions of the spaces of $G$-steerable kernels for each task.
$\gX( \gS \times \gA )$ refers to the space \textit{without} equivariant constraints, while $\gX( \gB )$ denotes the space with \textit{$G$-steerable equivariant constraints}.
We can see that the spaces of equivariant version are hugely reduced because of (1) smaller base space, as visualized in Figure~\ref{fig:demo-kernel}, and (2) constrained output matrix $\sR^{d \times d}$ with less free parameters.

In the empirical results, we show for multiple free particles ($N$-ball \texttt{PointMass} 3D), which generalize the results of single balls.
Compared to \texttt{Reacher} that has two connected links by a joint, multiple free particles can have much better saving.
Note that for \texttt{Reacher}, in implementation we convert the angles to unit vectors:
\begin{equation}
    \left(\theta_1, \theta_2, \dot{\theta}_1, \dot{\theta}_2, x_g-x_f, y_g-y_f\right) \Rightarrow\left(\cos \theta_1, \sin \theta_1, \cos \theta_2, \sin \theta_2, \dot{\theta}_1, \dot{\theta}_2, x_g-x_f, y_g-y_f\right).
\end{equation}

\subsubsection{Single Particle Control in the Plane}
We work out the single particle control problem in the plane. The goal is to move a particle to origin. The state space is the particle position, represented as a two-vector $\Vec{p}$, and the particle velocity represented as a two-vector $\Vec{v}$. The action space consists of the applied force $\hat{F}$, which is also a two-vector. The state space and action space are thus given by
\begin{align*}
    \gS = \rtwo \times \rtwo , \quad \quad \gA = \rtwo
\end{align*}
The group $SO(2)$ acts on the state and action space. Specifically, under a rotation $R \in SO(2)$, the state and action transform as
\begin{align*}
   & \text{State Transform: } ( \Vec{p} , \Vec{v} ) \rightarrow ( R \Vec{p} , R \Vec{v} )  \\
   & \text{Action Transform: } \Vec{F} \rightarrow R \Vec{F}
\end{align*}
The state space transforms in $ \rho_{\gS} =  \rho_{1} \oplus \rho_{1}$ and the action space transform in $\rho_{\gA} = \rho_{1}$. The set of orbits are then given by states and actions where the angles between the vectors $\Vec{p}$, $\Vec{v}$ and $\Vec{F}$ are fixed. 

The base space is given by $\base = \mathbb{R}^{+} \times \mathbb{R}^{4}$. Using Proposition E.6. in \cite{Lang_2020}, a basis for the steerable kernels of input type $(\rho_{1} , V_{1} )$ and output type $(\rho_{1} , V_{1} )$ is given by
\begin{align*}
K_{11}(x) = c_{1}\begin{bmatrix}
    1 & 0 \\
    0 & 1
\end{bmatrix} + c_{2}\begin{bmatrix}
    0 & -1 \\
    1 & 0
\end{bmatrix} + c_{3} \begin{bmatrix}
    \cos(2x) & \sin(2x) \\
    \sin(2x) & -\cos(2x)
\end{bmatrix} + c_{4} \begin{bmatrix}
    -\sin(2x) & \cos(2x) \\
    \cos(2x) & -\sin(2x)
\end{bmatrix} 
\end{align*}
where each $c_{j} \in \mathbb{R}$. This result was first derived in \cite{Weiler_2018}. In more compact notation, the matrix $K_{11}: S^{1} \to \mathbb{R}^{2\times 2}$ can be expanded as
\begin{align*}
\small
K_{11}(x) = \begin{bmatrix}
    c_{1} + c_{3} \cos(2x) - c_{4}\sin(2x) ,  & -c_{1} + c_{3}\sin(2x) - c_{4}\cos(2x) \\
    c_{2} + c_{3}\sin(2x) + c_{4}\cos(2x) , & c_{1} + c_{3}\sin(2x) - c_{4}\sin(2x) \\
\end{bmatrix}
\end{align*}
where each $c_{i} \in \mathbb{R}$. Then, using the form of the LQR matrices, we can write 
\begin{align*}
    A( p_{\downarrow} , x ) = \begin{bmatrix}
        K_{A}^{(1,1)}( p_{\downarrow},x) , & K_{A}^{(1,2)}( p_{\downarrow},x) \\
        K_{A}^{(2,1)}( p_{\downarrow},x) , & K_{A}^{(2,2)}( p_{\downarrow},x)
    \end{bmatrix}, \quad    B( p_{\downarrow} , x ) = \begin{bmatrix}
   K_{B}^{(1,1)}( p_{\downarrow},x) \\
   K_{B}^{(2,1)}( p_{\downarrow},x)
    \end{bmatrix}
\end{align*}
where each $K_{A}^{(i,j)}$ and $K_{B}^{(k,l)}$ take the form, $X\in \{A,B\}$, 
\begin{align*}
\small
K_{X}^{(i,j)}( p_{\downarrow} , x) = \begin{bmatrix}
    c_{1,X}^{(i,j)}(p_{\downarrow} ) + c_{3,X}^{(i,j)}(p_{\downarrow} )\cos(2x) - c_{4,X}^{(i,j)}(p_{\downarrow} )\sin(2x) ,  & -c_{1,X}^{(i,j)}(p_{\downarrow} ) + c_{3,X}^{(i,j)}(p_{\downarrow} )\sin(2x) - c_{4,X}^{(i,j)}(p_{\downarrow} )\cos(2x) \\
    c_{2,X}^{(i,j)}(p_{\downarrow} ) + c_{3,X}^{(i,j)}(p_{\downarrow} )\sin(2x) + c_{4,X}^{(i,j)}(p_{\downarrow} )\cos(2x) , & c_{1,X}^{(i,j)}(p_{\downarrow} ) + c_{3,X}^{(i,j)}(p_{\downarrow} )\sin(2x) - c_{4,X}^{(i,j)}(p_{\downarrow} )\sin(2x) \\
\end{bmatrix}
\end{align*}
the coefficients $c_{kA}^{(i,j)} : \mathbb{R}^{+} \times \mathbb{R}^{4} \rightarrow \mathbb{R} $ and $c_{kB}^{(i,j)} : \mathbb{R}^{+} \times \mathbb{R}^{4} \rightarrow \mathbb{R} $ are not constrained by symmetry and can be parameterized by a neural network. If we amalgamate each of the coefficients $c_{kA}^{(i,j)}$ and $c_{kB}^{(i,j)}$ into a single vector output $C$, the system dynamics can be learned by specifying 
\begin{align*}
  C : \mathbb{R}^{+} \times \mathbb{R}^{4} \rightarrow \mathbb{R}^{16 + 8}
\end{align*}
This should be contrasted with the non-equivarient case, where one needs to learn 
\begin{align*}
  A : \mathbb{R}^{6} \rightarrow \mathbb{R}^{16} \text{ and } B: \mathbb{R}^{6} \rightarrow \mathbb{R}^{8}
\end{align*}
The dimensional reduction in the equivarient vs non-equivarient case is infinite. This is analogous to \cite{wang_learning_2020} where equivarient methods can outperform non-equivarient methods by essentially an essentially infinite margin. In practice, due to discritization, this infinite gain is reduced to some large finite number.

\subsubsection{Single Particle Control in Space}

Let us consider the analogous single particle control problem in three-dimensional space. The state space is the particle position, represented as a three-vector $\Vec{p}$, and the particle velocity represented as a three-vector $\Vec{v}$. The action space consists of the applied force $\hat{F}$, which is also a three-vector. The state space and action space are thus given by
\begin{align*}
    \gS = \rthree \times \rthree , \quad \quad \gA = \rthree
\end{align*}
The group $SO(3)$ acts on the state and action space. Specifically, under a rotation $R \in SO(3)$, the state and action transform as
\begin{align*}
   & \text{State Transform: } ( \Vec{p} , \Vec{v} ) \rightarrow ( R \Vec{p} , R \Vec{v} )  \\
   & \text{Action Transform: } \Vec{F} \rightarrow R \Vec{F}
\end{align*}
Position and Velocity are both vectors the state space transforms in $ \rho_{\gS} =  \rho_{1} \oplus \rho_{1}$. Force is a vector quantity and the action space transform in $\rho_{\gA} = \rho_{1}$. The set of $SO(3)$-orbits are then given by states and actions where the angles between the vectors $\Vec{p}$, $\Vec{v}$ and $\Vec{F}$ are fixed. The base space is given by $\base = \mathbb{R}^{+} \times \mathbb{R}^{6}$. 

A complete characterization of $SO(3)$-kernels was first derived in \cite{Weiler_2018_II}. In the basis that diogonalizes the representation, the vectored kernel matrix for a $SO(3)$-steerable kernel of input type $(D^{1},W^{1} )$ and $(D^{1},W^{1} )$ can be written as
\begin{align*}
    \text{Vec}( K_{11}(x) ) = \begin{bmatrix}
        \Phi_{0}( ||x|| ) Y_{0}( \frac{x}{||x|| } ) \\ 
        \Phi_{1}( ||x|| ) Y_{1}( \frac{x}{||x|| }  )\\
        \Phi_{2}( ||x|| ) Y_{2}( \frac{x}{||x|| } ) \\ 
    \end{bmatrix}
\end{align*}
where each $Y_{\ell} : S^{2} \rightarrow \mathbb{R}^{(2\ell +1)} $ are spherical harmonics in vector form. Each $\Phi_{i}( ||x || ) : \mathbb{R}^{+} \rightarrow \mathbb{R} $ are a set of radial functions. Unvectorizing, we have that
\begin{align*}
    K_{11}(x) = \begin{bmatrix}
    \Phi_{0}( ||x|| ) Y^{0}_{0}( \frac{x}{||x|| } ) & \Phi_{1}( ||x|| ) Y^{1}_{1}( \frac{x}{||x|| } ) & \Phi_{2}( ||x|| ) Y^{2}_{2}( \frac{x}{||x|| } ) \\
    \Phi_{1}( ||x|| ) Y^{-1}_{1}( \frac{x}{||x|| } ) & \Phi_{1}( ||x|| ) Y^{0}_{1}( \frac{x}{||x|| } ) & \Phi_{2}( ||x|| ) Y^{1}_{2}( \frac{x}{||x|| } ) \\
    \Phi_{1}( ||x|| ) Y^{-2}_{2}( \frac{x}{||x|| } ) & \Phi_{2}( ||x|| ) Y^{-1}_{2}( \frac{x}{||x|| } ) & \Phi_{2}( ||x|| ) Y^{0}_{2}( \frac{x}{||x|| } ) \\
    \end{bmatrix}
\end{align*}

Now, using the results of $G$-steerable kernel constraints, the most general LQR matrices can be written in the form
\begin{align*}
    A( p_{\downarrow} , x ) = \begin{bmatrix}
        K_{A}^{(1,1)}( p_{\downarrow},x) , & K_{A}^{(1,2)}( p_{\downarrow},x) \\
        K_{A}^{(2,1)}( p_{\downarrow},x) , & K_{A}^{(2,2)}( p_{\downarrow},x)
    \end{bmatrix}, \quad    B( p_{\downarrow} , x ) = \begin{bmatrix}
   K_{B}^{(1,1)}( p_{\downarrow},x) \\
   K_{B}^{(2,1)}( p_{\downarrow},x)
    \end{bmatrix}
\end{align*}
where each $K_{A}^{(i,j)}$ and $K_{B}^{(k,l)}$ take the form, $X\in \{A,B\}$, 
\begin{align*}
\small
K^{ij}_{X}(x) = \begin{bmatrix}
    \Phi^{ij}_{0}( ||x|| ) Y^{0}_{0}( \frac{x}{||x|| } ) & \Phi^{ij}_{1}( ||x|| ) Y^{1}_{1}( \frac{x}{||x|| } ) & \Phi^{ij}_{2}( ||x|| ) Y^{0}_{2}( \frac{x}{||x|| } ) \\
    \Phi^{ij}_{1}( ||x|| ) Y^{-1}_{1}( \frac{x}{||x|| } ) & \Phi^{ij}_{2}( ||x|| ) Y^{-2}_{-2}( \frac{x}{||x|| } ) & \Phi^{ij}_{2}( ||x|| ) Y^{1}_{2}( \frac{x}{||x|| } ) \\
    \Phi^{ij}_{1}( ||x|| ) Y^{1}_{3}( \frac{x}{||x|| } ) & \Phi^{ij}_{2}( ||x|| ) Y^{-1}_{2}( \frac{x}{||x|| } ) & \Phi^{ij}_{2}( ||x|| ) Y^{2}_{2}( \frac{x}{||x|| } ) \\
    \end{bmatrix}
\end{align*}
the coefficients $\Phi_{kA}^{(i,j)} : \mathbb{R}^{+} \times \mathbb{R}^{6} \rightarrow \mathbb{R} $ and $\Phi_{kB}^{(i,j)} : \mathbb{R}^{+} \times \mathbb{R}^{6} \rightarrow \mathbb{R} $ are not constrained by symmetry and can be parameterized by a neural network. If we amalgamate each of the coefficients $\Phi_{kB}^{(i,j)}$ and $\Phi_{kB}^{(i,j)}$ into a single vector output $\Phi$, the system dynamics can be learned by specifying 
\begin{align*}
  \Phi : \mathbb{R}^{+} \times \mathbb{R}^{6} \rightarrow \mathbb{R}^{12 + 6}
\end{align*}
This should be contrasted with the non-equivarient case where one needs to learn a function from $\mathbb{R}^{3} \times \mathbb{R}^{6} \rightarrow \mathbb{R}^{12 + 6}$. By utilizing symmetry, we are able to significantly reduce the domain of the function spaces.

\section{Implementation Details and Additional Evaluation}
\label{sec:add_results}

\subsection{Implementation Details: Equivariant TD-MPC}

We mostly follow the implementation of TD-MPC \citep{hansen_temporal_2022}.
The training of TD-MPC is end-to-end, i.e., it produces trajectories with a learned dynamics and reward model and predicts the values and optimal actions for those states.
It closely resembles MuZero \citep{schrittwieser_mastering_2019} while uses MPPI (Model Predictive Path Integral \citep{williams_model_2015,williams_information_2017}) for continuous actions instead of MCTS (Monte-Carlo tree search) for discrete actions.
It inherits the drawbacks from MuZero - the dynamics model is trained only from reward signals and may collapse or experience instability on sparse-reward tasks.
This is also the case for the tasks we use: \texttt{PointMass} and \texttt{Reacher} and their variants, where the objectives are to reach a goal position.

\subsection{Experimental Details}

We implement $G$-equivariant MLP using \texttt{escnn} \citep{weiler_general_2021} for policy, value, transition, and reward network, with 2D and 3D discrete groups.
For all MLPs, we use two layers with $512$ hidden units. The hidden dimension is set to be $48$ for non-equivariant version, and the equivariant version is to keep the same number of free parameters, or \texttt{sqrt} strategy.

For example, for $D_8$ group, \texttt{sqrt} strategy (to keep same free parameters) has number of hidden units divided by $\sqrt{|D_8|} = \sqrt{16}=4$.
The other strategy is to make equivariant networks' input and output be compatible with non-equivariant ones: \textit{\texttt{linear}} strategy, which keeps same input/output dimensions (number of hidden units divided by $|D_8| = 16$).

The hidden space uses \textit{regular} representation, which is common for discrete equivariant network \citep{cohen_group_2016,weiler_general_2021,zhao_integrating_2022}.

\subsection{Additional Results}

\begin{figure*}[t]
\centering
\subfigure{
\includegraphics[width=0.5\linewidth]{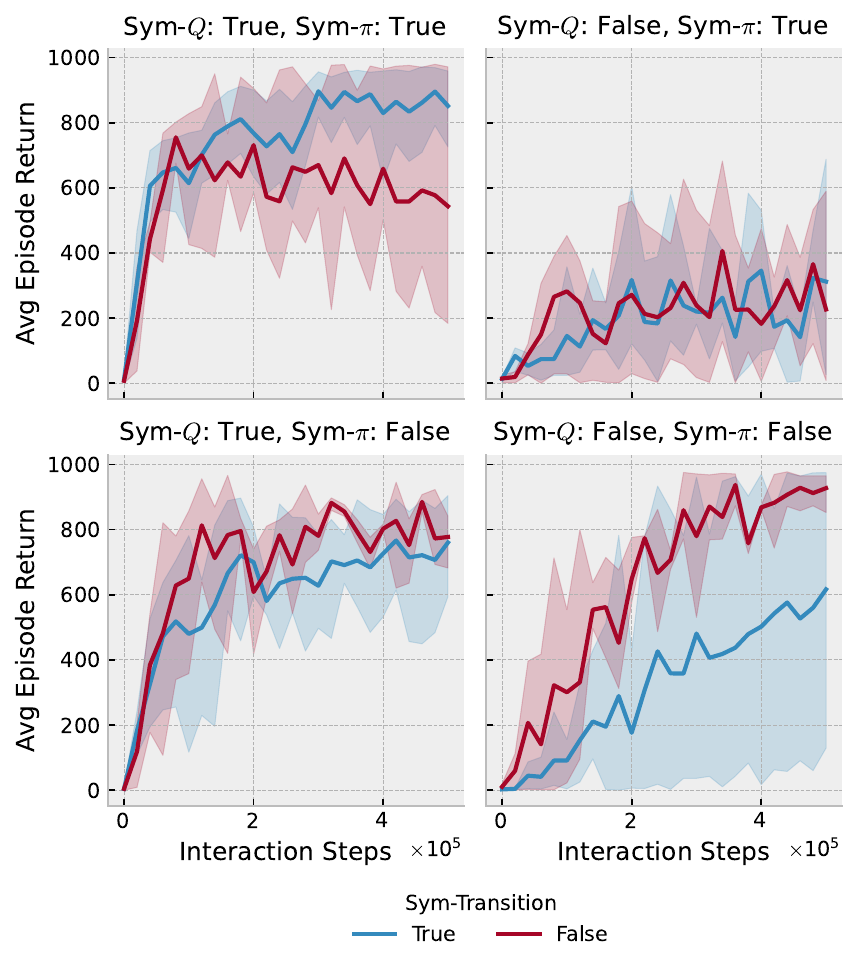}
}
\centering
\caption{
Ablation study on equivariant components, using \texttt{Reacher} Hard with $D_8$ symmetry group.
}
\label{fig:ablation-equiv-componets}
\end{figure*}

\paragraph{Ablation on equivariant components.}

Recall that we have several equivariant components in equivariant TD-MPC:
\begin{align}
f_\theta &: \gS \times \gA \to \gS :& \rho_\gS(g) \cdot f_\theta(\vs_{t}, \va_{t}) &=  f_\theta(\rho_\gS(g) \cdot \vs_{t}, \rho_\gA(g) \cdot \va_{t}) \\
R_\theta &: \gS \times \gA \to \sR :& {R}_\theta(\vs_{t}, \va_{t} ) &=  {R}_\theta(\rho_\gS(g) \cdot \vs_{t}, \rho_\gA(g) \cdot \va_{t}) \\
Q_\theta &: \gS \times \gA \to \sR :& {Q}_\theta(\vs_{t}, \va_{t} ) &=  {Q}_\theta(\rho_\gS(g) \cdot \vs_{t}, \rho_\gA(g) \cdot \va_{t}) \\
\pi_\theta &: \gS \to \gA  :& \rho_\gA(g) \cdot {\pi}_\theta(\cdot \mid \vs_{t}) &=  {\pi}_\theta(\cdot \mid \rho_\gS(g) \cdot \vs_{t})
\end{align}

We experiment to enable and disable each of them: (1) transition network: dynamics $f$ and reward $R$, (2) value network: $Q$, and (3) policy network $\pi$.
Note that to make equivariant and non-equivariant components compatible, we need to make sure the input and output dimensions match.

We show the results on \texttt{Reacher} Hard with $D_8$ symmetry group in Fig~\ref{fig:ablation-equiv-componets}.
Instead of using \texttt{sqrt} strategy (to keep same free parameters, number of hidden units divided by $\sqrt{|D_8|} = \sqrt{16}=4$) on specifying the number of hidden units, we use \textit{\texttt{linear}} strategy that keeps same input/output dimensions (number of hidden units divided by $|D_8| = 16$). Thus, the performance of fully non-equivariant model and fully equivariant model are not directly comparable, because the number of free parameters in fully equivariant one is much smaller.

The results show the relative importance of value, policy, and transition.
It shows the most important equivariant component is $Q$-value network. It is reasonable because it has been used intensively in predicting into the future, where generalization and training efficiency are very important and benefit from equivariance.

\begin{figure*}[t]
\centering
\subfigure{
\includegraphics[width=1.\linewidth]{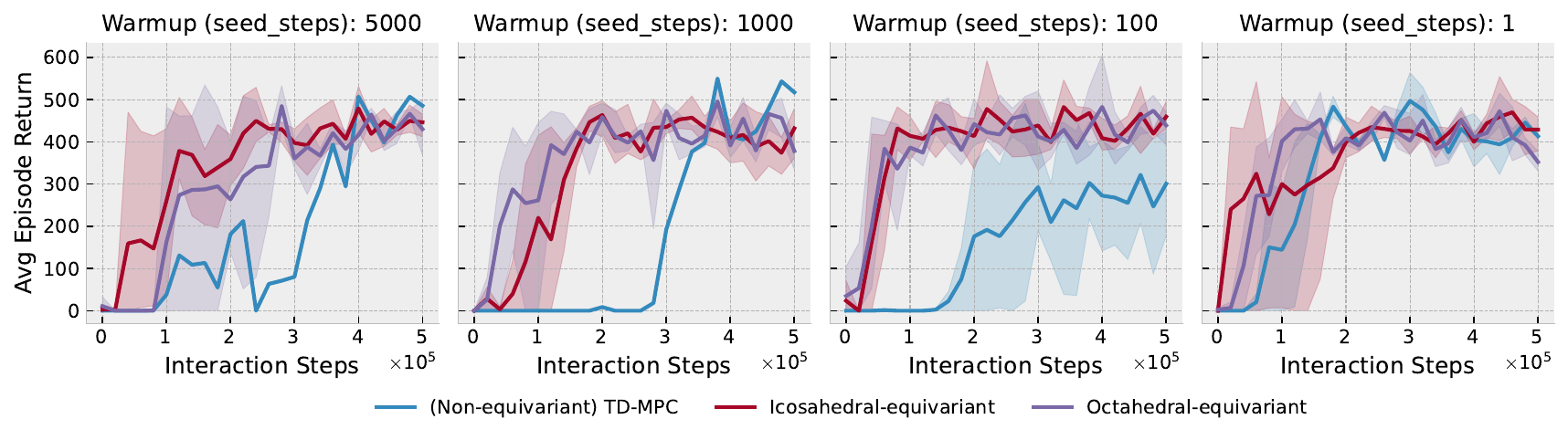}
}
\centering
\caption{
Ablation study on number of warmup episodes on \texttt{PointMass} \textit{3D with small target}.
}
\label{fig:ablation-warmup}
\end{figure*}

\paragraph{Hyperparameter of amount of warmup.}
We experiment different number of warmup episodes, called \texttt{seed steps} in TD-MPC hyperparameter.
We find this is a critical hyperparameter for (non-equivariant) TD-MPC.
One possible reason is that TD-MPC highly relies on joint training and may collapse when the transition model is stuck at some local minima. This warmup hyperparameter controls how many episodes TD-MPC collects before starting actual training.

We test using different numbers on \texttt{PointMass} \textit{3D with small target}.
The results are shown in Figure~\ref{fig:ablation-warmup}, which demonstrate that our equivariant version is robust under all choices of warmup episodes, even with little to none warmup.
The non-equivariant TD-MPC is very sensitive to the choice of warmup number.

\begin{figure*}[t]
\centering
\subfigure{
\includegraphics[width=0.5\linewidth]{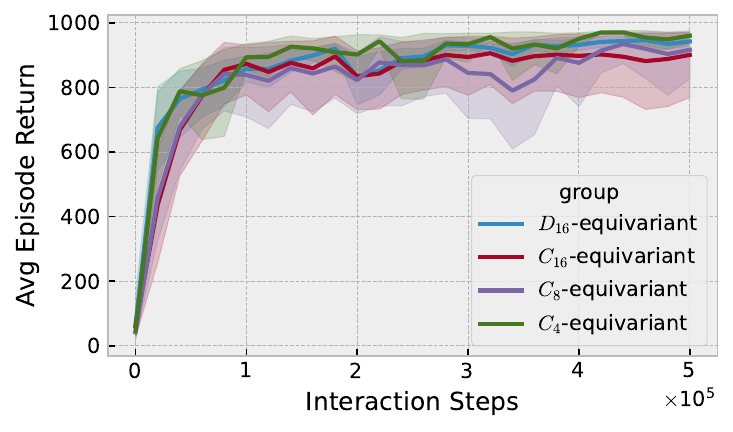}
}
\centering
\caption{
Ablation study on symmetry group on \texttt{Reacher} Hard.
}
\label{fig:ablation-symmetry}
\end{figure*}

\paragraph{Ablation on symmetry groups.}

We also do ablation study on the choice of discrete subgroups.
We run experiments on \texttt{Reacher} Hard to compare 2D discrete rotation/dihedral groups: $C_4, C_8, C_{16}, D_{16}$, using $1$ warmup episode.

The results are shown in Fig~\ref{fig:ablation-symmetry}. We find using groups larger than $C_8$ does not bring additional improvement on this specific task, \texttt{Reacher} Hard.
In the main paper, we thus use $D_4, D_8$ to balance the performance and computation time and memory use.

\paragraph{Comparing reference frames and state features.}

This experiment studies the balance between reference frames and the choice of state features.
In the theory section, we emphasize that kinematic constraints introduce local reference frames.

Here, we study a specific example: \texttt{Reacher} (Easy and Hard).
The second joint has angle $\theta_2$ and angular velocity $\dot \theta_2$ relative to the first link.

For \textit{local} reference frame version, we use
\begin{equation}
    \left(\theta_1, \theta_2, \dot{\theta}_1, \dot{\theta}_2, x_g-x_f, y_g-y_f\right) \Rightarrow\left(\cos \theta_1, \sin \theta_1, \cos \theta_2, \sin \theta_2, \dot{\theta}_1, \dot{\theta}_2, x_g-x_f, y_g-y_f\right)
\end{equation}
Thus, $\cos \theta_1, \sin \theta_1$ is transformed under standard representation $\rho_1$ and $\cos \theta_2, \sin \theta_2$ is transformed under trivial representation $\rho_0 \oplus \rho_0$.

For the \textit{global} reference frame version, we compute the global location of the end-effector (tip) by adding the location of the first joint.
Thus, the global position is transformed also under standard representation now $\rho_1$.

\begin{figure*}[t]
\centering
\subfigure{
\includegraphics[width=0.4\linewidth]{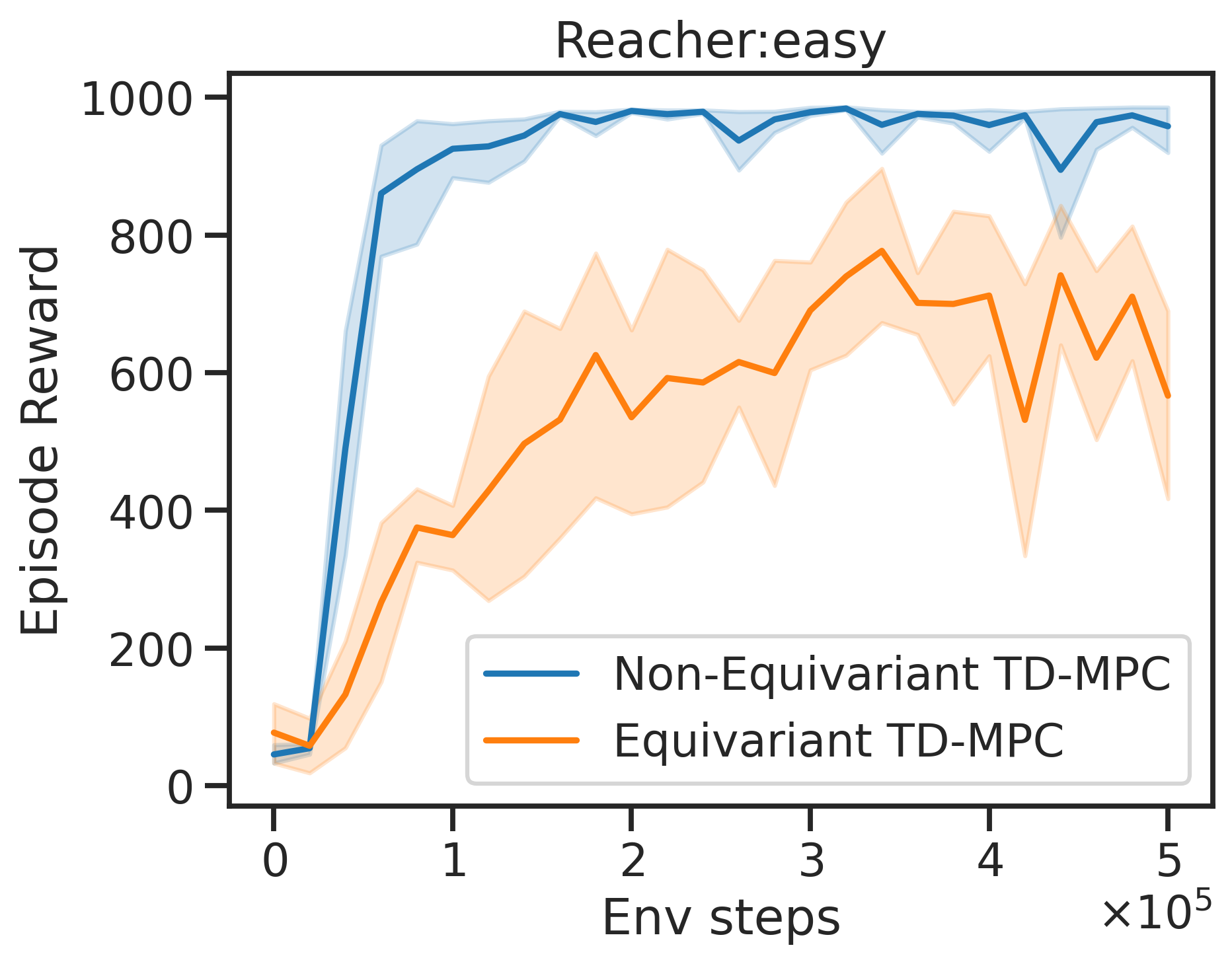}
}
\subfigure{
\includegraphics[width=0.4\linewidth]{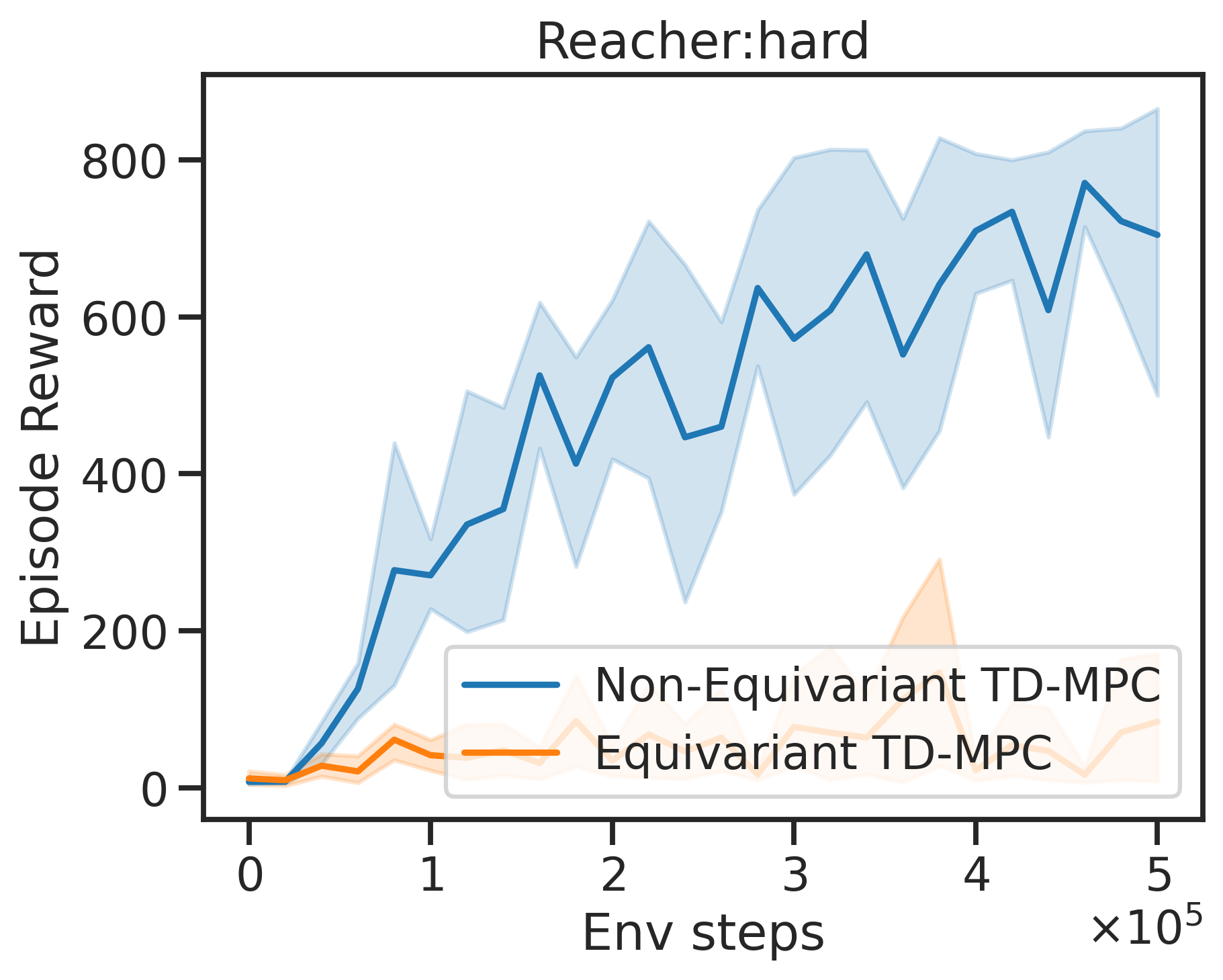}
}
\centering
\caption{
Results for global reference frame on \texttt{Reacher}.
}
\label{fig:env-reward-curves-global}
\end{figure*}

We show the results in Figure~\ref{fig:env-reward-curves-global}.
Evaluation reward curves for non-equivariant and equivariant TD-MPC over $5$ runs using global frames. Error bars denote $95\%$ confidence intervals. Non-equivariant TD-MPC outperforms equivariant TD-MPC.
Surprisingly, we find using global reference frame where the second joint is associated with standard representation (equivariant feature, instead of invariant feature) brings much worse results, compared to the local frame version in the main paper.
One possibility is that it is more important to encode kinematic constraints (e.g., the length of the second link is preserved in $\cos \theta_2, \sin \theta_2$), compared to using equivariant feature.

\section{Additional Mathematical Background}

We provide additional mathematical background on representation theory and its relation with our theoretical results.

\subsection{Mathematical Exposition: Principal $G$-Bundles and LQR}

Fiber bundles were introduced in the 1930s as a natural extension of the concept of tangent spaces in differential geometry \citep{Milnor_1974}. Conceptually, fiber bundles attach additional information at each point on some underlying manifold. The connection between equivarient machine learning and the theory of fiber-bundles was first noted in \citep{Cohen_2016_II,Weiler_2018}. For continuous state-action manifolds, the dynamics can be viewed as some additional structure that depends on the state action manifold location. The results presented in the main text can be understood within the context of fiber bundle theory.

\subsubsection{Fiber Bundle}

We briefly comment on how Equivarient LQR can be understood in terms of fiber bundles. For a full exposition on the theory of fiber-bundles, please see \citep{Karstoft_1992,Husemöller_2013}. 

Formally, a smooth fiber bundle is specified by $(E, B , F , \pi)$ where $E$ and $B$ are smooth manifolds and $F$ is a vector space. $E$ is called the \emph{total space} and $B$ is called the \emph{base space}. The vector space $V$ is called the \emph{fiber}. The map $\pi : E \rightarrow B$ is a smooth subjection called the projection map. The projection $\pi$ must satisfy a trivialization condition \citep{Husemöller_2013}.

Sections are generalization of vector fields. A smooth section $s$ of a fiber bundle is a smooth map $s: B \rightarrow E$ such that
\begin{align*}
\forall x \in B, \quad    \pi(  s(x) ) = x
\end{align*}
so that a section does not change the base point. One can think of a section $s(x)$ as a vector with origin at point $x\in B$.

\subsubsection{ Principal Bundle }

Intuitively, a $G$-Principal Bundle is fiber bundle with an additional symmetry on the total space. Let $G$ be a Lie group. Then, a smooth $G$-Principal Bundle is smooth fiber bundle that has continuous $G$ action on the total space $G \times E \rightarrow E$ which preserves the fibers of $E$ so that
\begin{align*}
\forall x \in E, \enspace  \forall g \in G, \quad \pi( g \cdot x ) =  \pi(x) 
\end{align*}

Sections of $G$-Principal Bundles are defined analogously to the vector bundle case. Specifically, a section $s$ is a map $s: B \rightarrow E$ such that
\begin{align*}
\forall x \in B, \quad    \pi(  s(x) ) = x
\end{align*}
Note that if $s: B \rightarrow E$ is a section then $g \cdot s : B \rightarrow E $ is another section as
\begin{align*}
  \forall x \in B, \enspace \forall g \in G \quad    \pi( g \cdot s(x) ) = \pi(  s(x) ) = x 
\end{align*}
Thus, on $G$-Principal bundles we can speak of $G$-families of sections $s_{g} : G \times B \rightarrow E$ defined as $ s_{g}(x) = g \cdot s(x) $. It is thus natural to consider equivalence classes of sections. These equivalence classes correspond to flows related by $G$ action.

More formally, the LQR system in consideration is a $G$-bundle with total space given by $E = \mathcal{S} \times \mathcal{A}$ and base space $ B =  \base$ given by $G$-orbit representatives. The canonical projection $\pi : \mathcal{S} \times \mathcal{A} \rightarrow B $ is the map unto $G$-orbit representatives which satisfies 
\begin{align*}
 \forall g \in G, \quad \Pi( g \cdot x )  =   \Pi(x) 
\end{align*}
The LQR dynamical matrices $A$ and $B$ can be viewed as sections of the $G$-bundle. By utilizing equivarience methods, we only need to learn a section on the base space $B$ as opposed to vector fields on the total space $E$. This corresponds to learning equivalence classes of sections, instead of individual sections.

\end{document}